\documentclass{article} 
\usepackage{iclr2025_conference,times}

\usepackage{amsmath,amsfonts,bm}

\def\eqref#1{equation~\ref{#1}}

\def\1{\bm{1}}

\DeclareMathAlphabet{\mathsfit}{\encodingdefault}{\sfdefault}{m}{sl}
\SetMathAlphabet{\mathsfit}{bold}{\encodingdefault}{\sfdefault}{bx}{n}

\usepackage{url}
\usepackage{arydshln}
\usepackage{epigraph}
\usepackage{graphicx}
\usepackage{wrapfig}

\usepackage[dvipsnames]{xcolor}
\definecolor{softblue}{RGB}{100, 130, 180}  
\usepackage[colorlinks=true, citecolor=softblue]{hyperref}
\usepackage{hwemoji}

\newcommand{\name}{PhysToolBench}

\title{\includegraphics[scale=0.086]{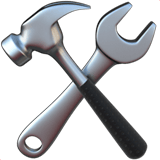}~\name: Benchmarking Physical\\ Tool Understanding for MLLMs}

\iclrfinalcopy

\author{
\textbf{Zixin Zhang}$^{1,}$\footnotemark[1]~, 
\textbf{Kanghao Chen}$^{1,}$\footnotemark[1]~, 
\textbf{Xingwang Lin}$^{3}$, 
\textbf{Lutao Jiang}$^{1}$, 
\textbf{Xu Zheng}$^{1}$, 
\textbf{Yuanhuiyi Lyu}$^{1}$, \\
~\textbf{Litao Guo}$^{1}$, 
\textbf{Yinchuan Li}$^{4}$, 
\textbf{Ying-Cong Chen}$^{1,2,}$\footnotemark[2] \\
~$^{1}$HKUST(GZ) $^{2}$HKUST
$^{3}$Beihang University $^{4}$Knowin
}

\begin{document}

\maketitle
\begin{figure}[h]
    \centering
    \vspace{-9mm}
    \includegraphics[width=0.95\linewidth]{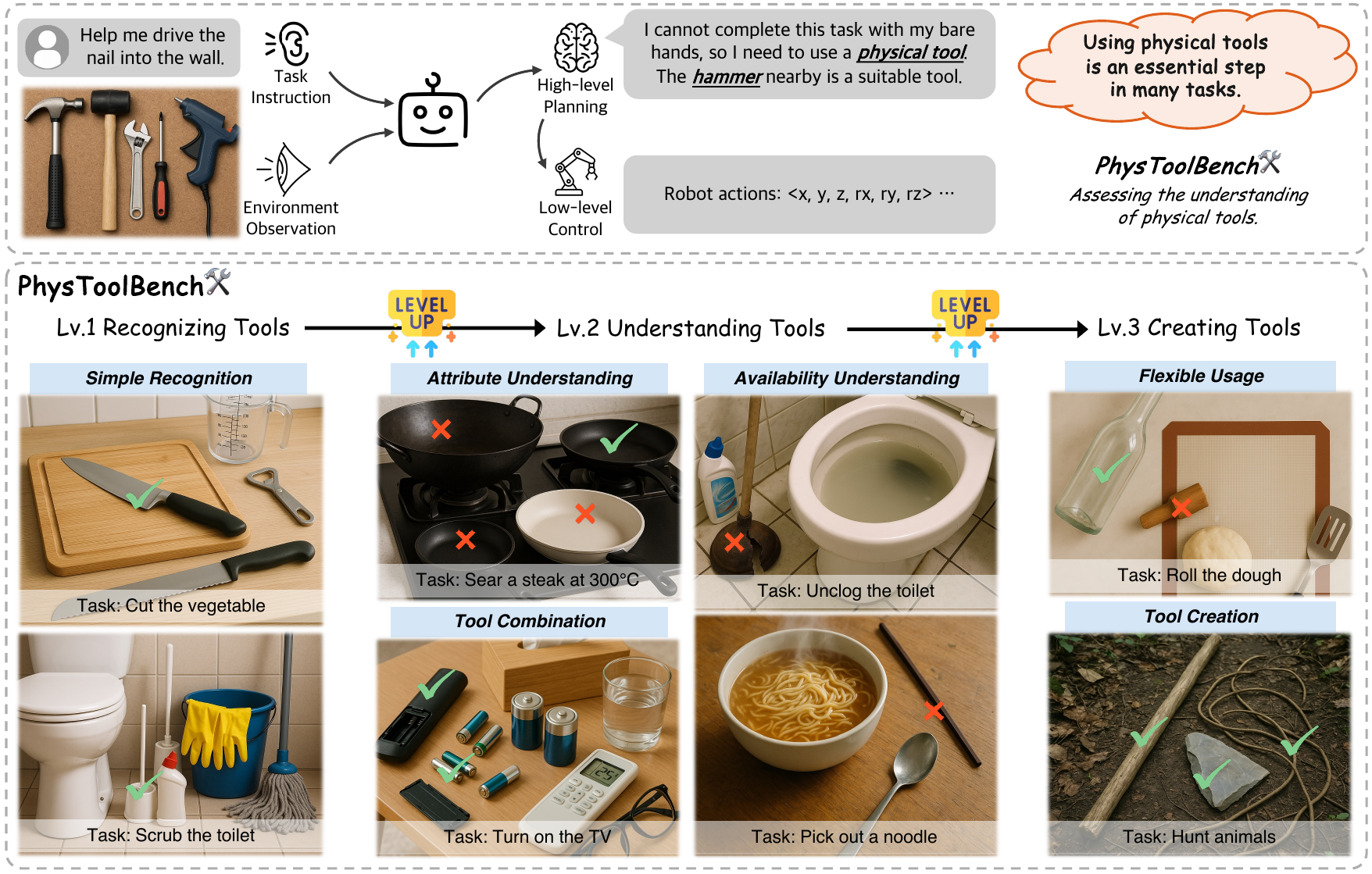}
    \vspace{-4mm}
    \caption{For an Embodied Agent, using physical tools is crucial in many tasks. The understanding of physical tools significantly impacts the task's success rate and execution efficiency{~(Top)}. \name{~(Bottom)} systematically evaluates the understanding of physical tools of multimodal LLMs. The benchmark is designed with three progressive levels of difficulty and employs a Visual Question Answering (VQA) format. Notice that in the actual benchmark, tools in the images are numerically labeled, and images here are for illustrative purposes only.}
    \label{fig:teaser}
\end{figure}

\begin{abstract}
The ability to use, understand, and create tools is a hallmark of human intelligence, enabling sophisticated interaction with the physical world. For any general-purpose intelligent agent to achieve true versatility, it must also master these fundamental skills. While modern Multimodal Large Language Models (MLLMs) leverage their extensive common knowledge for high-level planning in embodied AI and in downstream Vision-Language-Action (VLA) models, the extent of their true understanding of physical tools remains unquantified. To bridge this gap, we present \textbf{\name}, the first benchmark dedicated to evaluating the comprehension of physical tools by MLLMs.
Our benchmark is structured as a Visual Question Answering (VQA) dataset comprising over 1,000 image-text pairs. It assesses capabilities across three distinct difficulty levels:
\textit{1)} Tool Recognition: Requiring the recognition of a tool's primary function.
\textit{2)} Tool Understanding: Testing the ability to grasp the underlying principles of a tool's operation.
\textit{3)} Tool Creation: Challenging the model to fashion a new tool from surrounding objects when conventional options are unavailable.
Our comprehensive evaluation of 32 MLLMs—spanning proprietary, open-source, specialized embodied, and backbones in VLAs—reveals a significant deficiency in the tool understanding. Furthermore, we provide an in-depth analysis and propose preliminary solutions.
Code and dataset are publicly available at \href{https://github.com/EnVision-Research/PhysToolBench}{PhysToolBench Repository}.
\end{abstract}

\section{Introduction}

\setlength{\epigraphwidth}{0.7\textwidth}

\vspace{-3mm}
\epigraph{\textit{Man is a tool-using animal. Without tools, he is nothing; with tools, he is all.}}{---Thomas Carlyle}
\vspace{-3mm}

A key factor in humanity's success throughout natural evolution is the ability to create and utilize a vast array of tools to enhance survival and prosperity. With the advancement of technology, humans continuously reshape the physical world, inventing diverse instruments to extend the boundaries of their capabilities. For an embodied intelligent agent designed to complete physical tasks, the use of tools is a prerequisite for achieving success and efficiency. For instance, as illustrated in Fig.~\ref{fig:teaser}~(Top), a robot must use a hammer to drive a nail into a wall—a task it cannot accomplish with its bare manipulators. Arguably, a profound understanding of physical tools is a fundamental precondition for Artificial General Intelligence (AGI).

Multimodal Large Language Models (MLLMs) ~\citep{bai2025qwen2, gpt-4o, openai2025gpt5, claude-3-7-card, gemini2.5}, which can process inputs from both vision and language modalities, have acquired substantial common-sense knowledge from being trained on massive datasets. They show great promise for evolving into AGI and have been the focus of numerous studies for deployment in robotics. Some studies employ MLLMs as high-level planners~\citep{yuan2025embodied, team2025robobrain, driess2023palm}, while others utilize them for low-level control as the backbone of Vision-Language-Action (VLA) models~\citep{black2024pi_0, kim2024openvla, wen2025dexvla,blackP05VisionLanguageActionModel}. In either case, interaction with the physical world is fundamental, which inevitably involves the use of physical tools. Although some research has demonstrated that MLLMs possess a preliminary understanding of tools~\citep{gao2025vlmgineer, tang2025mimicfunc, trupin2025dynamic}, the true depth of physical tool comprehension remains largely unexplored.
 
Based on these considerations, we propose \textbf{\textit{\name}}, a benchmark for evaluating an agent's understanding of physical tools. To the best of our knowledge, this is the first benchmark specifically designed for this purpose.
To evaluate an agent's practical capabilities, we designed a Visual Question Answering (VQA) benchmark that simulates a robotic workflow. Presented with a task and an image of objects, the agent must select the appropriate tool(s). As shown in Fig.~\ref{fig:teaser}~(Bottom), the benchmark features three difficulty levels to progressively assess the agent's depth of understanding:
\textbf{\textit{1) Easy (Recognizing Tools)}}. This fundamental level assesses whether an agent can identify a conventional tool and its primary function.
\textbf{\textit{2) Medium (Understanding Tools)}}. This intermediate level probes the agent's comprehension through three distinct challenges: optimal tool selection from functionally similar options, selection of all tools required for a multi-tool task, and assessment of a tool's operational viability based on its physical state.
\textbf{\textit{3) Hard (Creating Tools)}}. This advanced level evaluates an agent's inventive capabilities. Faced with a task and no standard tools, the agent must fashion a solution by repurposing or combining available objects, which requires an understanding of the physical principles underlying the required tool.

We evaluate the performance of 32 MLLMs on \name, spanning four distinct classes: general-purpose proprietary MLLMs, general-purpose open-source MLLMs, MLLMs tailored for embodied AI, and those functioning as backbones in VLAs.
The results demonstrate a clear performance ceiling, with even the most advanced proprietary models scoring no higher than 63\%, revealing a profound disparity with human proficiency in tool understanding (over 90\%). Furthermore, our analysis uncovers several critical weaknesses in current MLLMs: (1) a failure of small MLLMs, including those within VLA models, to exhibit an emergent ability of tool understanding; (2) a long-tail distribution issue in recognizing and understanding a wide array of tools; (3) a tendency to hallucinate tool affordances and their availability; and (4) inadequate visual reasoning skills. We further propose a "vision-centric reasoning" framework to bolster the visual reasoning of MLLM agents. 
We hope our work will inspire future research on physical tool understanding. 
\section{Related Works}
\subsection{MLLM and its Application in Embodied AI}
Recent years have witnessed remarkable advancements in Multimodal Large Language Models (MLLMs). Building on the significant success of Large Language Models (LLMs), these models effectively process visual information by leveraging modality alignment techniques~\citep{li2022blip, clip}. Typically, a visual encoder and a connector are employed to link visual data to the LLM, enabling reasoning at the language level and granting Vision-Language Models (VLMs) sophisticated image comprehension capabilities. To date, numerous impressive MLLMs have emerged, including proprietary models~\citep{gpt-4o, openai2025gpt5, gemini2.5, claude-3-7-card, grok4_modelcard_2025}, as well as open-source alternatives~\citep{bai2025qwen2, zhu2025internvl3, wang2025internvl3-5, wu2024deepseek, beyer2024paligemma, wu2024deepseek, lu2025ovis2}. These models have demonstrated powerful visual understanding across a diverse range of tasks.

Beyond general-purpose domains, MLLMs are also finding significant applications in embodied intelligence. On one hand, they are being utilized as the high-level "brain" for task planning in embodied agents, as exemplified by PaLM-E~\citep{driess2023palm}, RoboBrain~\citep{team2025robobrain}, and Embodied-R1~\citep{yuan2025embodied}. On the other hand, research has also capitalized on the inherent common-sense knowledge within MLLMs. By adding an action head and fine-tuning on robotic data, they can be transformed into end-to-end Vision-Language-Action (VLA) models capable of directly outputting robot actions. Notable examples of this approach include $\pi_0$~\citep{black2024pi_0}, $\pi_{0.5}$~\citep{blackP05VisionLanguageActionModel}, and OpenVLA~\citep{kim2024openvla}.

\subsection{Physical Tool Use in Embodied AI}
These advancements in foundation models have empowered robots with the ability to perform fundamental tasks when these models are embodied. For instance, embodied models such as $\pi_0$, $\pi_{0.5}$, and OpenVLA can successfully accomplish basic household chores like folding clothes and tidying desktops. However, while these tasks can be efficiently completed using only the robot's own manipulators, many higher-level, real-world tasks are difficult and even impossible to achieve with robot manipulators alone. Consequently, teaching robots how to use tools to effectively accomplish complex objectives is of critical importance.

Initial research has begun to explore endowing robots with tool-using capabilities. For example, VLMgineer~\citep{gao2025vlmgineer} employs a VLM agent to assist robots in crafting simple tools to complete tasks. Similarly, ~\cite{trupin2025dynamic} leverages vision foundation models to enable tool use during task planning. MimicFunc~\citep{tang2025mimicfunc} establishes an imitation learning framework that allows robots to learn tool manipulation by observing human demonstration videos. Leveraging the common-sense knowledge inherent in MLLMs, these approaches have demonstrated a rudimentary ability to use physical tools. Nevertheless, the depth of physical tool understanding that their "brain"—the MLLM—possesses remains largely unexplored. The primary motivation for our work is to clarify this question by creating a benchmark designed specifically to evaluate the understanding of physical tools within Multimodal Large Language Models.

\subsection{Related Benchmarks}
For Large Language Models (LLMs), a multitude of benchmarks~\citep{tool1, tool2, tool3, tool4, tool5} have been developed to evaluate their ability to utilize digital tools, such as search engines, translation services, booking systems, etc.. These benchmarks have catalyzed the rapid development of modern LLM Agents, equipping them with the capability to invoke external APIs to accomplish complex tasks. However, a significant gap exists when it comes to physical tools, as there is currently no corresponding benchmark for MLLMs. We argue that such a benchmark is crucial for advancing MLLMs toward becoming true Embodied Agents capable of meaningful interaction with the physical world.

Among existing benchmarks for MLLMs, A4Bench~\citep{wang2025affordance} is the most relevant to our research. It operates in a VQA format, presenting an image of a tool and asking the MLLM to identify its function from a set of multiple-choice options. While this can, to some extent, reflect the MLLM's understanding of object affordances, we contend that this question-answering format lacks practical applicability. Our work, therefore, aims to establish a more application-oriented evaluation. We provide the MLLM with a specific task requirement and an image containing several tools, compelling it to answer the question based on the observation. This approach more rigorously assesses whether the MLLM can apply genuine knowledge and reasoning to find the optimal tool, rather than merely relying on the rote memorization of tool-function associations.

\section{The \name}

\subsection{Overview}
\begin{figure}[t]
    \centering
    \includegraphics[width=1\linewidth]{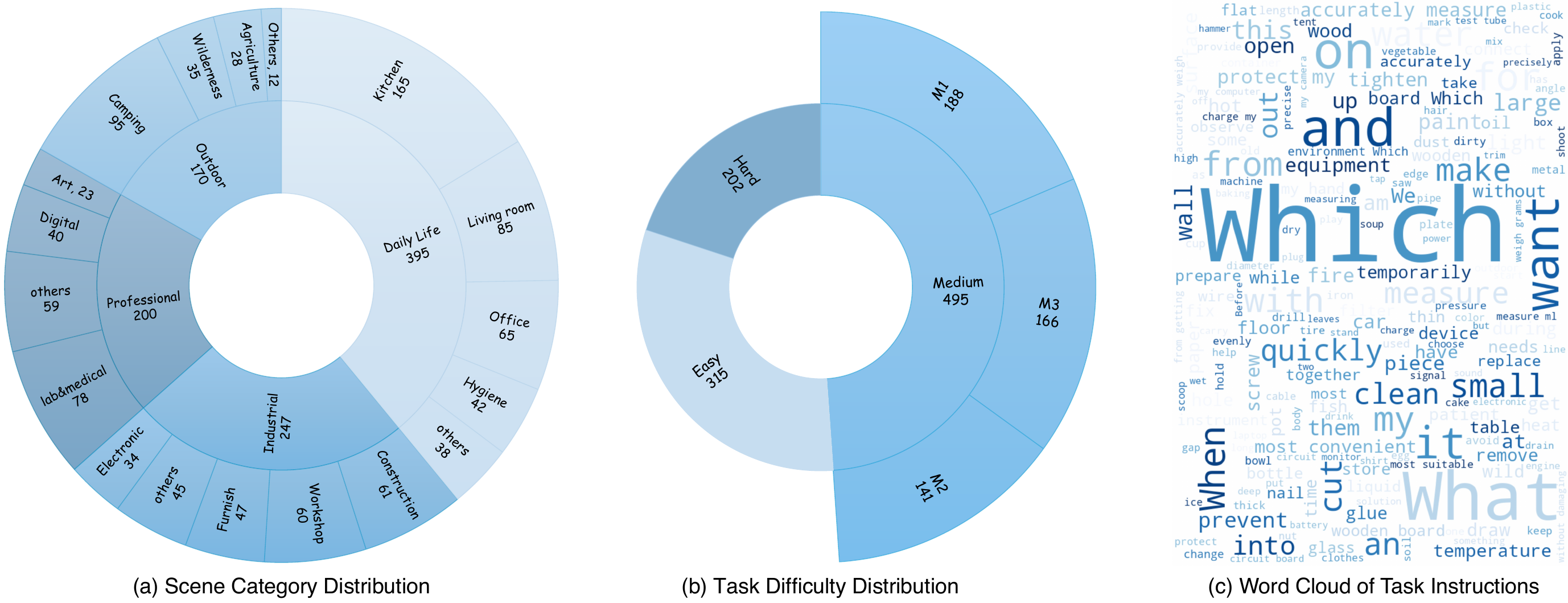}
    \vspace{-6mm}
    \caption{Statistics of \name. (a) is the distribution of the category. (b) is the distribution of the difficulty level. (c) is the word cloud of the task description given to MLLMs.}
    \label{fig:statistics}
\end{figure}

\name is a VQA benchmark comprising over 1,000 text-image pairs designed to evaluate an MLLM's understanding of physical tools. Each pair consists of a text prompt outlining a specific task and a corresponding $1024\times1024$ image displaying several numerically labeled tools and objects. A core design constraint is that the MLLM is explicitly instructed that the items depicted in the image are the only available things, simulating a realistic robotics scenario with limited resources. The MLLM's objective is to analyze the task and visual information, then output the numerical label(s) of the required tool(s), or "None" if no suitable tool is available. \name spans four major domains: Daily Life, Industrial, Outdoor Activities, Professional Settings, and three difficulty levels: Easy, Medium, Hard. Detailed statistics are shown in Fig.~\ref{fig:statistics}.

\subsection{Design Principles}
To progressively evaluate the depth of an MLLM's understanding, we designed \name with three distinct difficulty levels: Easy, Medium, and Hard, each demanding a more profound comprehension of tool properties and functionality.

The \textbf{Easy} level assesses fundamental tool recognition. Questions are answerable with basic tool identification and common-sense knowledge. Task prompts are straightforward, and the image always contains a tool whose primary function directly matches the task. For example, to "cut vegetables," the image will include a kitchen knife.
The \textbf{Medium} level requires a deeper understanding of tools, necessitating reasoning based on specific task constraints. This tier is subdivided into three challenges: 1) \textit{M.1. Attribute Understanding}, requiring comprehension of a tool's specific attributes (e.g., selecting a cast-iron skillet for its high heat tolerance); 2) \textit{M.2. Tool Combination}, evaluating the ability to combine tools to unlock new affordances (e.g., inserting batteries into a remote); and 3) \textit{M.3. Availability Understanding}, testing the recognition of non-functional tools (e.g., identifying a cracked plunger as unusable).
The \textbf{Hard} level assesses higher-order reasoning and creativity. The model must work backwards from task requirements to innovatively utilize surrounding objects. For instance, if tasked to "tighten a flat-head screw" without a suitable screwdriver, the MLLM must identify that a coin can serve as a substitute.

We propose these difficulty levels as a tiered evaluation standard. The 'Easy' score serves as a prerequisite for basic tool-use planning, 'Medium' benchmarks potential in complex scenarios, and 'Hard' presents a forward-looking challenge for AGI research.

\subsection{Dataset Collection Process}
\label{sec:dataset}

The collection of test samples for \name was conducted in three phases to ensure quality.
\textit{\textbf{Phase 1: Conceptualization.}} Human experts designed task-scene pairs, consisting of a task requirement and a detailed scene description, meticulously aligning each scenario with our Easy, Medium, and Hard difficulty criteria.
\textit{\textbf{Phase 2: Image Generation.}} Scene descriptions were transformed into visual images primarily using GPT-4o-image~\citep{openai2025gpt4oimage}(approximately 90\%), a process closely supervised by human experts who vetted for quality and realism. For complex objects that the generative model struggled with, we resorted to physical staging and photography(approximately 10\%).
\textit{\textbf{Phase 3: Annotation and Verification.}} Experts used a custom software tool to apply numerical labels to objects in each image. The entire dataset then underwent a final, thorough review and revision by a separate team to verify its integrity and ensure reliability. More details are provided in Appendix.~\ref{appendix:more}.

\section{Experiments on \name}
\begin{figure}[t]
    \centering
    \includegraphics[width=1\linewidth]{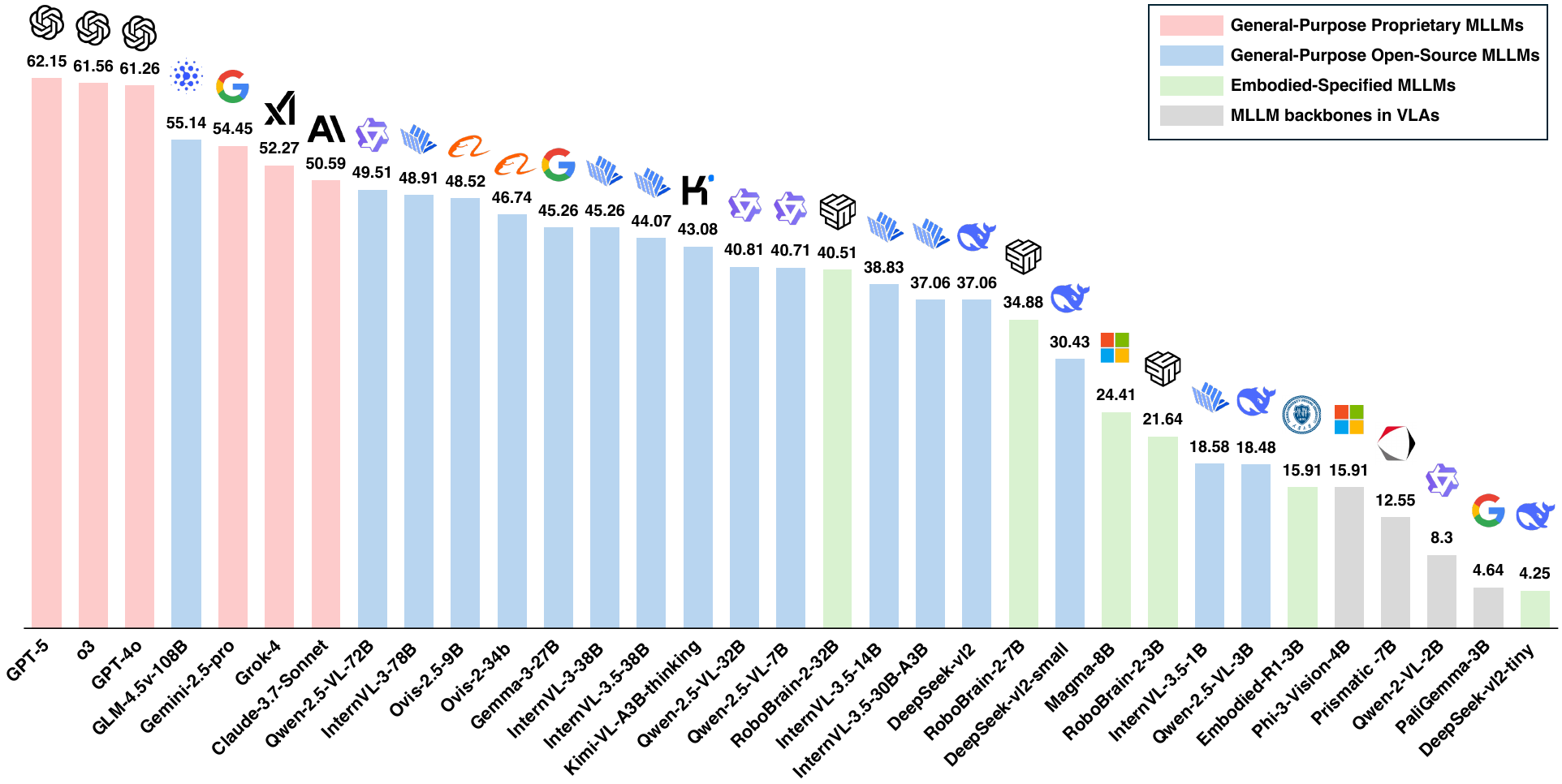}
    \vspace{-7mm}
    \caption{MLLM Leaderboard on our \name, ranked by overall performance.}
    \label{fig:leaderboard}
\end{figure}

\subsection{Benchmark Candidates}
We conducted a comprehensive evaluation across four distinct categories of state-of-the-art Multimodal Large Language Models (MLLMs), encompassing \textbf{\textit{32 models}} in total:
\noindent\textbf{\textit{a) General-Purpose Proprietary MLLMs:}} GPT-5 (2025-08-17)~\citep{openai2025gpt5}, o3 (2025-04-16)~\citep{openai2025o3_o4mini}, ChatGPT-4o-latest (2025-01-29)~\citep{gpt-4o}, Claude-3-7-Sonnet-thinking~\citep{claude-3-7-card}, Gemini-2.5-pro (2025-05-06)~\citep{gemini2.5}, Grok-4~\citep{grok4_modelcard_2025}. 
\noindent\textbf{\textit{b) General-Purpose Open-Source MLLMs:}} Qwen-2.5-VL-72B-Instruct ~\citep{bai2025qwen2}, Qwen-2.5-VL-32B-Instruct, Qwen-2.5-VL-7B-Instruct, Qwen-2.5-VL-3B-Instruct, InternVL-3.5-38B~\citep{wang2025internvl3-5}, InternVL-3.5-30B-A3B, InternVL-3.5-14B, InternVL-3.5-1B, InternVL-3-78B~\citep{zhu2025internvl3}, InternVL-3-38B, GLM-4.5V-108B~\citep{vteam2025glm45vglm41vthinkingversatilemultimodal}, Ovis-2-34B~\citep{lu2024ovis}, Ovis-2.5-9B~\citep{lu2025ovis2}, DeepSeek-VL-2~\citep{wu2024deepseek}, DeepSeek-VL-2-small, DeepSeek-VL-2-tiny, Kimi-VL-A3B-thinking-2506~\citep{team2025kimi}. 
\noindent\textbf{\textit{c) Embodied-Specific MLLMs:}} RoboBrain-2-32B~\citep{team2025robobrain}, RoboBrain-2-7B, RoboBrain-2-3B, Embodied-R1-3B~\citep{yuan2025embodied}, Magma-8B~\citep{yang2025magma}
\noindent\textbf{\textit{d) MLLM Backbones of Vision-Language-Action (VLA) models:}} Prismatic-7B~\citep{karamcheti2024prismatic} in OpenVLA~\citep{kim2024openvla}, PaliGemma-3B~\citep{beyer2024paligemma} in $\pi_0$~\citep{black2024pi_0}, Qwen-2-VL-2B~\citep{Qwen2VL} in DexVLA~\citep{wen2025dexvla}, Phi-3-Vision-4B~\citep{Phi3} in TraceVLA~\citep{zheng2024tracevla}.

The first category of proprietary models was evaluated via their respective APIs. For the latter three categories, the models were downloaded and deployed locally for testing.
To ensure a fair comparison, we used a consistent text prompt for all models. The system prompt was designed to encourage a Chain-of-Thought~\citep{cot}, explicitly asking the models to reason before providing their final answer. The only exception was for models that feature a native, built-in "thinking" mode, in which case we allowed them to utilize their default inference process without modification. We also recruited 5 human participants as testers to serve as a reference.

\begin{table*}[t]\scriptsize
    \centering
    \renewcommand\arraystretch{1.1}
    \setlength{\tabcolsep}{2pt}
    \vspace{-3mm}
    \caption{Benchmark results on the \name. For each difficulty level and scene category, the best performance was marked in \textbf{bold} and the second best was marked \underline{underline}. *Prismatic-7B achieves an unusually high score on the Medium-M3 difficulty. Upon inspecting its reasoning process, we discovered that the model does not generate sound reasoning but instead exhibits a strong tendency to output "None" in all case.}
    \begin{tabular}{l|ccccc|cccc|c}
    \hline
    \hline
         \textbf{Categories} & \multicolumn{5}{c|}{\textbf{Difficulty Level}} & \multicolumn{4}{c|}{\textbf{Scene Category}}  &{\textbf{Overall$\uparrow$}}\\ \cdashline{1-10}
        {\textbf{MLLM}}  & {\textit{Easy$\uparrow$}}& {\textit{m1$\uparrow$}} & {\textit{m2$\uparrow$}} & {\textit{m3$\uparrow$}} & {\textit{Hard$\uparrow$}} & \textit{Professional$\uparrow$}  & \textit{Industrial$\uparrow$} & \textit{Outdoor$\uparrow$} & {\textit{Daily$\uparrow$}} \\\hline

\textsc{Human(Best)} & 96.19\% & 93.61\% & 90.78\% & 93.97\% & 89.10\% & 87.5\% & 93.52\% & 91.17\% & 96.71\% & 93.19\% \\
\textsc{Human(Worst)} & 91.74\% & 87.77\% & 85.11\% & 90.36\% & 81.68\% & 80.5\% & 85.02\% & 87.65\% & 93.42\% & 87.85\% \\
\hline
\multicolumn{11}{l}{\textbf{General-Purpose Proprietary MLLMs:}} \\ \hline
\textsc{Gemini-2.5-pro} & 78.10\% & 48.40\% & 46.10\% & \underline{45.78\%} & 36.14\% & 58.5\% & 61.54\% & 46.47\% & 51.39\% & 54.45\% \\
\textsc{o3} & \textbf{93.02\%} & \underline{67.02\%} & 46.81\% & 22.89\% & \textbf{49.50\%} & \underline{64.0\%} & \textbf{68.02\%} & \textbf{61.18\%} & 56.46\% & \underline{61.56\%} \\
\textsc{gpt-4o} & 86.03\% & \textbf{70.74\%} & \underline{48.23\%} & 35.54\% & 44.06\% & 62.5\% & 63.97\% & \underline{59.41\%} & \textbf{59.75\%} & 61.26\% \\
\textsc{gpt-5} & 90.16\% & 63.83\% & \textbf{50.35\%} & 36.75\% & \underline{46.04\%} & \textbf{67.5\%} & \underline{66.8\%} & 58.82\% & \underline{57.97\%} & \textbf{62.15\%} \\
\textsc{grok-4} & 73.65\% & 46.28\% & 30.50\% & \textbf{52.41\%} & 39.60\% & 50.5\% & 59.92\% & 43.53\% & 52.15\% & 52.27\% \\
\textsc{claude-3-7-sonnet-thinking} & 74.60\% & 58.51\% & 35.46\% & 27.11\% & 35.64\% & 53.5\% & 55.87\% & 45.88\% & 47.85\% & 50.59\% \\
\hline
\multicolumn{11}{l}{\textbf{General-Purpose Open-Source MLLMs:}} \\ \hline
\textsc{Qwen-2.5-VL-72B} & 75.56\% & 55.85\% & 35.46\% & 31.93\% & 27.23\% & 51.5\% & 55.47\% & 44.71\% & 46.84\% & 49.51\% \\
\textsc{Qwen-2.5-VL-32B} & 67.62\% & 43.09\% & 30.5\% & 22.29\% & 19.31\% & 42.0\% & 49.39\% & 37.06\% & 36.46\% & 40.81\% \\
\textsc{Qwen-2.5-VL-7B} & 71.43\% & 51.6\% & 20.57\% & 21.08\% & 12.87\% & 44.0\% & 49.39\% & 38.24\% & 34.68\% & 40.71\% \\
\textsc{Qwen-2.5-VL-3B} & 36.51\% & 10.64\% & 6.38\% & 13.86\% & 9.9\% & 21.5\% & 21.46\% & 15.88\% & 16.2\% & 18.48\% \\
\textsc{GLM-4.5v-108B} & \underline{90.48\%} & 65.43\% & 36.88\% & 16.27\% & 35.15\% & 62.5\% & 59.92\% & 56.47\% & 47.85\% & 55.14\% \\
\textsc{Gemma-3-27B} & 68.57\% & 57.45\% & 31.91\% & 19.88\% & 27.72\% & 50.0\% & 48.99\% & 42.94\% & 41.52\% & 45.26\% \\
\textsc{InternVL-3.5-38B} & 70.79\% & 50.53\% & 29.08\% & 18.67\% & 27.72\% & 51.0\% & 49.8\% & 37.65\% & 39.75\% & 44.07\% \\
\textsc{InternVL-3.5-30B-A3B} & 66.03\% & 37.77\% & 20.57\% & 15.06\% & 20.79\% & 41.0\% & 43.32\% & 31.18\% & 33.67\% & 37.06\% \\
\textsc{InternVL-3.5-14B} & 66.03\% & 40.43\% & 21.99\% & 21.08\% & 21.29\% & 44.0\% & 44.94\% & 31.76\% & 35.44\% & 38.83\% \\
\textsc{InternVL-3.5-1B} & 38.73\% & 19.68\% & 4.26\% & 3.61\% & 8.42\% & 22.5\% & 18.22\% & 20.0\% & 16.2\% & 18.58\% \\
\textsc{InternVL-3-78B} & 79.05\% & 53.72\% & 39.01\% & 21.08\% & 27.23\% & 52.0\% & 56.28\% & 42.94\% & 45.32\% & 48.91\% \\
\textsc{InternVL-3-38B} & 77.78\% & 44.68\% & 31.91\% & 16.87\% & 27.72\% & 51.0\% & 53.04\% & 41.18\% & 39.24\% & 45.26\% \\
\textsc{Ovis-2.5-9B} & 80.63\% & 55.85\% & 42.55\% & 17.47\% & 21.29\% & 57.0\% & 56.28\% & 44.12\% & 41.27\% & 48.52\% \\
\textsc{Ovis-2-34B} & 83.17\% & 45.21\% & 35.46\% & 15.66\% & 24.75\% & 56.5\% & 52.23\% & 40.0\% & 41.27\% & 46.74\% \\
\textsc{DeepSeek-vl2-27B} & 71.75\% & 39.89\% & 19.86\% & 6.63\% & 17.33\% & 44.0\% & 42.91\% & 35.88\% & 30.38\% & 37.06\% \\
\textsc{DeepSeek-vl2-small-16B} & 64.44\% & 28.19\% & 10.64\% & 10.24\% & 9.9\% & 36.0\% & 37.65\% & 25.88\% & 25.06\% & 30.43\% \\
\textsc{DeepSeek-vl2-tiny-3B} & 7.62\% & 2.13\% & 2.84\% & 4.22\% & 1.98\% & 7.0\% & 5.26\% & 2.94\% & 2.78\% & 4.25\% \\
\textsc{Kimi-VL-30B-A3B-thinking} & 79.05\% & 45.21\% & 31.21\% & 18.67\% & 13.37\% & 46.5\% & 48.58\% & 40.59\% & 38.99\% & 43.08\% \\
\hline
\multicolumn{11}{l}{\textbf{Embodied-Specified MLLMs:}} \\ \hline
\textsc{RoboBrain-2-32B} & 75.87\% & 49.47\% & 19.86\% & 6.63\% & 19.31\% & 48.5\% & 47.37\% & 39.41\% & 32.66\% & 40.51\% \\
\textsc{RoboBrain-2-7B} & 66.03\% & 44.68\% & 13.48\% & 10.84\% & 11.88\% & 36.5\% & 41.7\% & 34.71\% & 29.87\% & 34.88\% \\
\textsc{RoboBrain-2-3B} & 46.35\% & 18.62\% & 3.55\% & 11.45\% & 6.93\% & 25.5\% & 28.74\% & 18.24\% & 16.71\% & 21.64\% \\
\textsc{Embodied-R1-3B} & 38.41\% & 6.38\% & 4.96\% & 4.22\% & 6.93\% & 23.0\% & 20.24\% & 11.76\% & 11.39\% & 15.91\% \\
\textsc{Magma-8B} & 46.35\% & 29.26\% & 0\% & 3.01\% & 20.3\% & 19.0\% & 29.55\% & 25.88\% & 23.29\% & 24.41\% \\
\hline
\multicolumn{11}{l}{\textbf{MLLM backbones in VLAs:}} \\ \hline
\textsc{PaliGemma-3B} & 7.94\% & 10.11\% & 0\% & 0\% & 1.49\% & 6.0\% & 4.86\% & 4.12\% & 4.05\% & 4.64\% \\
\textsc{Phi-3-Vision-4B} & 33.97\% & 12.77\% & 4.26\% & 3.01\% & 9.41\% & 20.5\% & 19.43\% & 11.18\% & 13.42\% & 15.91\% \\
\textsc{Qwen-2-VL-2B} & 19.37\% & 1.6\% & 0.71\% & 7.83\% & 2.97\% & 7.0\% & 9.31\% & 4.12\% & 10.13\% & 8.3\% \\
\textsc{Prismatic-7B} & 6.98\% & 4.26\% & 1.42\% & *56.02\% & 0.99\% & 11.0\% & 13.77\% & 8.24\% & 14.43\% & 12.55\% \\
\hline
\hline
\end{tabular}

    \label{tab:result}
    \vspace{-3mm}
\end{table*}

\subsection{Overall Results}
As shown in Tab.~\ref{tab:result}, MLLMs generally underperform, with most scoring below 60\%—a result far inferior to human performance, which consistently achieves at least 87.85\% overall accuracy. This indicates that contemporary MLLMs have a superficial understanding of tool usage.
Among the models evaluated, proprietary general-purpose MLLMs performed best. The OpenAI series (o3, gpt-4o, and gpt-5) all exceeded the 60\% threshold, with gpt-5 leading the group. Open-source general-purpose MLLMs followed, typically scoring above 40\%. GLM-4.5V was a notable exception, achieving 55.14\% and outperforming not only its open-source peers but also some proprietary models, highlighting its significant potential. Embodied-specific MLLMs demonstrated some capability but lagged behind the general-purpose open-source models. Lastly, MLLM backbones within VLA frameworks exhibited the weakest performance, likely due to their limited number of parameters. An overall leaderboard of MLLMs is shown in Fig.~\ref{fig:leaderboard}. We provide a set of complete VQA results in Appendix.~\ref{vqa}.

\subsection{Findings on \name}

\begin{figure}[t]
    \centering
    \begin{minipage}[t]{0.51\linewidth}
        \centering
        \includegraphics[width=\linewidth]{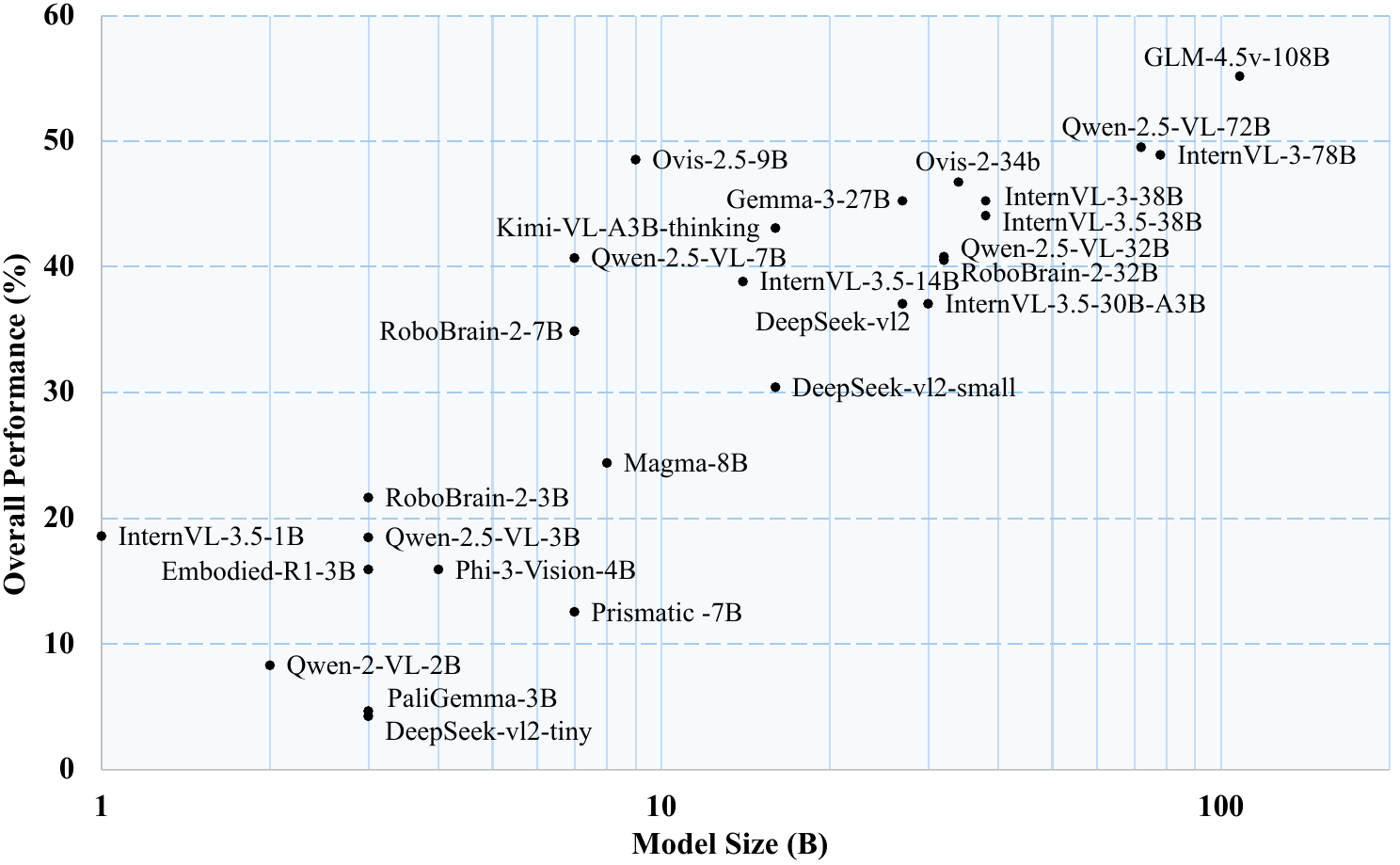}
        \vspace{-7.3mm}
        \caption{\textbf{Overall performance v.s. model size} for open-source MLLMs. A significant correlation is observed between performance and model size.}
        \label{fig:size}
    \end{minipage}
    \hfill
    \begin{minipage}[t]{0.46\linewidth}
        \vspace{-43.5mm}
        \centering
        \includegraphics[width=\linewidth]{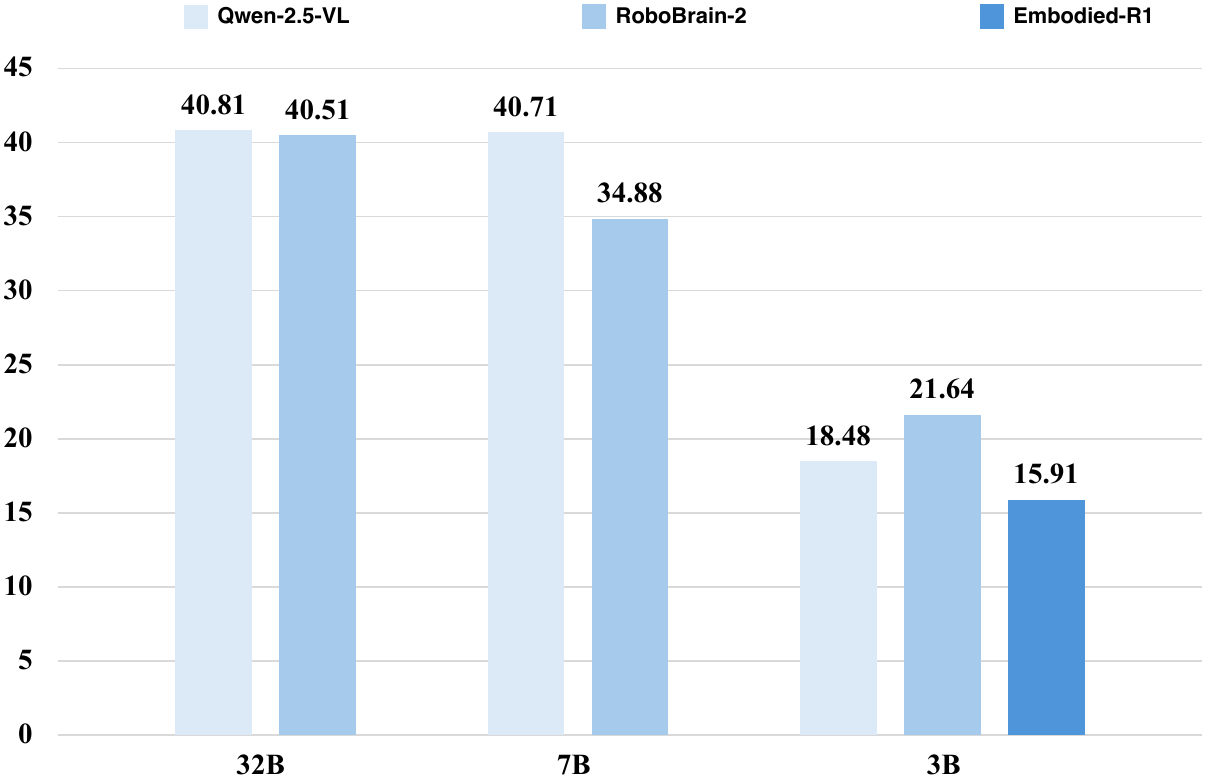}
        \vspace{-5mm}
        \caption{Performance comparison between the embodied models and their base model.}
        \label{fig:embodied}
    \end{minipage}
    \vspace{-3mm}
\end{figure}

\noindent \textbf{\textit{F.1. A foundational ability to understand tools emerges in large models with sufficient scale.}}
As shown in Fig.~\ref{fig:size}, our evaluation of numerous open-source models reveals that there's a significant correlation between the understanding of physical tool and the size of the model. 
Furthermore, for the easy difficulty setting in Tab.~\ref{tab:result}, we also observe that a foundational understanding of tool usage emerges once a model reaches a certain scale, which we preliminarily identify as approximately 10 billion parameters.
Most models exceeding this 10B threshold achieve an accuracy of 60-70\% on easy-level tasks. In contrast, performance drops significantly for smaller models; those with fewer than 5B parameters generally score below 50\% on easy tasks and have an overall accuracy below 25\%. Consequently, we recommend selecting MLLMs with more than 10 billion parameters for applications in embodied intelligence.

\begin{figure}[t]
    \centering
    \includegraphics[width=1\linewidth]{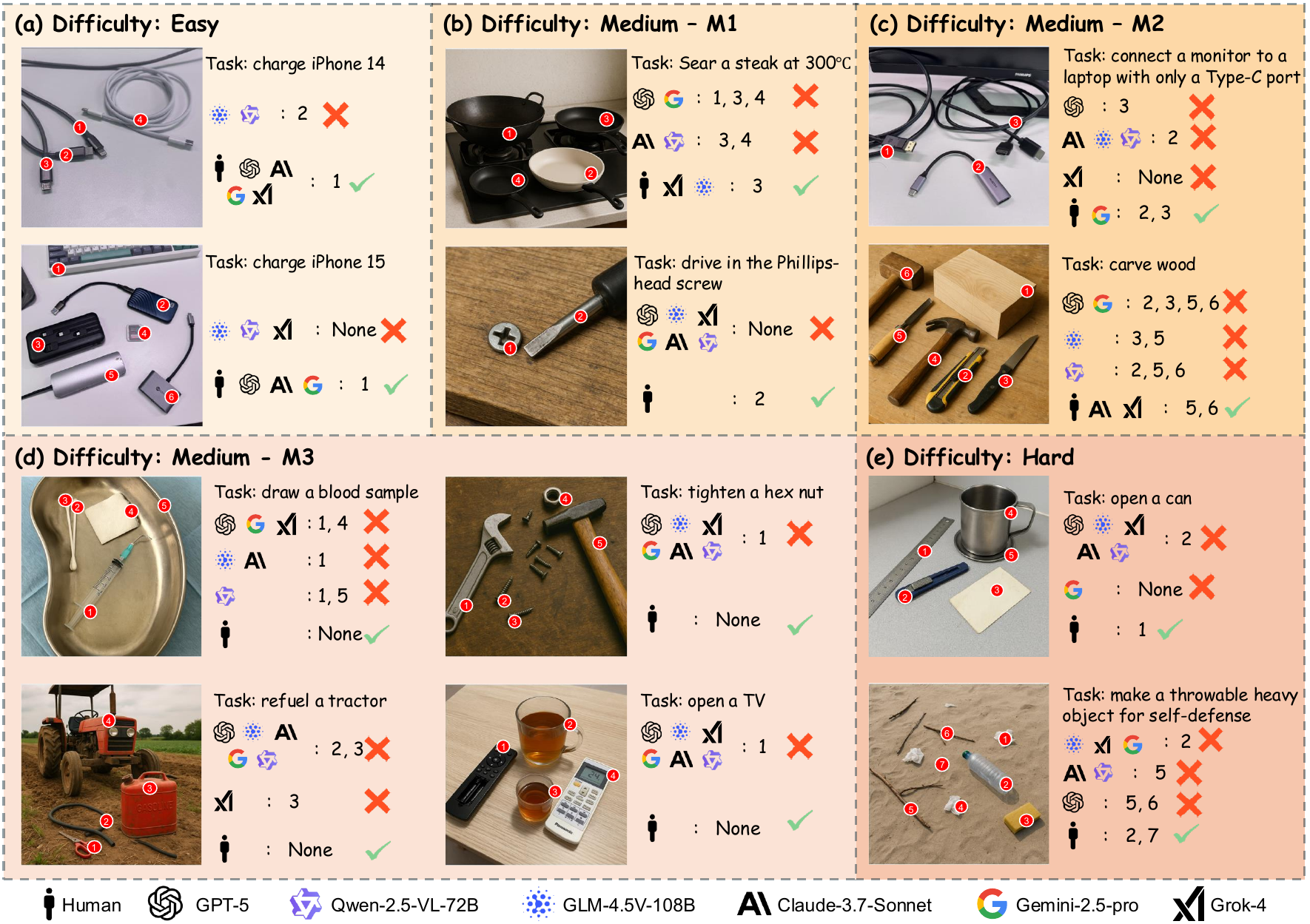}
    \vspace{-6mm}
    \caption{Some results of \name. We showcase illustrative examples for each difficulty level, along with the answers from several top-tier models and human participants. Note that the markers are intentionally enlarged for visualization purposes.}
    \label{fig:case}
    \vspace{-3mm}
\end{figure}

\noindent \textbf{\textit{F.2. A long-tail problem persists in tool recognition and understanding, even for the most advanced MLLMs.}} 
Although top-tier MLLMs are proficient at identifying common objects, their performance diminishes for less common items, creating a long-tail effect. A notable finding is the models' pronounced weakness in the subcategory of digital products. They frequently fail to distinguish between visually similar items, such as HDMI versus DP cables and Type-C versus Lightning charging ports. This deficiency is widespread in open-source models, where even the highly capable GLM-4V shows errors in basic recognition, as in Fig.~\ref{fig:case}.~(a). Closed-source models offer a marginal improvement but still demonstrate only a shallow comprehension. As an example in Fig.~\ref{fig:case}.~(c), most top-tier proprietary models do not grasp the functional requirement that a monitor must be connected to a laptop using an HDMI cable and an adapter if the laptop only has a Type-C port.

\noindent \textbf{\textit{F.3. Embodied-specific MLLMs show no significant advantage on \name.}} 
Models specifically fine-tuned for embodied tasks, such as RoboBrain2 and Embodied-R1, do not exhibit a notable performance improvement on our benchmark. RoboBrain2's parameters were initialized from Qwen2.5VL and subsequently fine-tuned on a combination of general vision and robotic datasets. Nevertheless, as shown in Fig.~\ref{fig:embodied}, its 32B, 7B variants all performed slightly below their Qwen2.5VL backbone of equivalent scale. A similar trend was observed with Embodied-R1-3B, which, despite being fine-tuned from Qwen-2.5-VL-3B, also achieved a marginally lower score than the original model. These findings indicate that the fine-tuning process did not confer an enhanced understanding of tools. We hypothesize that current robotic datasets may require more high-quality data centered on tool comprehension to advance these models' physical tool understanding.

\noindent \textbf{\textit{F.4. MLLMs exhibit a critical deficiency in comprehending tool availability, failing to grasp the fundamental principles of their utility.}} 
The M3 difficulty tier of our benchmark was specifically designed to probe this issue by incorporating simple "traps": presenting the correct tool for a task but in a damaged or non-functional state. 
Counter-intuitively, as shown in Tab.~\ref{tab:result}, models found this task more difficult than the "Hard" tier, which requires complex reasoning for tool creation.
For instance, in the selected four cases in Fig.~\ref{fig:case}.~(d), none of the MLLMs could identify when the tools are unavailable.  This outcome strongly suggests that the models' comprehension of tools is shallow and relies on surface-level "common sense" associations rather than a robust understanding of their core functionality, leading them to hallucinate the tool's usability.

The implications of this hallucination for embodied agents are severe. An agent that cannot recognize a tool as non-functional may attempt to use it, resulting in mission failure and significant safety hazards—for instance, fueling a tractor with gasoline, drawing a blood sample with a damaged syringe. We contend that addressing this issue is critical for advancing embodied AI.

\noindent \textbf{\textit{F.5. The MLLM backbones in current VLAs are extremely weak.}}
Our evaluation revealed that the MLLM backbones of the contemporary VLA models exhibit exceptionally poor performance on \name, with overall scores universally below 15\%. This result calls into question the prevailing assumption that VLAs can effectively inherit "common sense" from their base MLLM and then achieve generalization through fine-tuning on robotic action datasets. Our findings suggest that the foundational "common sense" of these MLLMs is profoundly insufficient for general-purpose intelligence. We posit that this fundamental limitation cannot be rectified through fine-tuning on robotic datasets of a modest scale. 
Consequently, we conclude that advancing the VLA paradigm will require a two-pronged approach: first, leveraging significantly larger and more capable MLLMs as backbones; and second, a substantial expansion in the size and diversity of robotic action datasets.

\noindent \textbf{\textit{F.6. Reasoning ability is important and useful, but still insufficient.}}
The capability for reasoning is crucial. In our experiments, we evaluated a subset of models under two conditions: one with Chain-of-Thought (CoT) prompting and one without. As shown in Tab.~\ref{tab:reasoning}, the models prompted with CoT demonstrated significantly higher accuracy. Furthermore, models that are natively optimized for reasoning exhibit superior performance. For instance, GLM-4.5V, the top-performing open-source model, was trained with a strong emphasis on reasoning. Its training regimen included not only Supervised Fine-Tuning (SFT) on high-quality CoT datasets but also reinforcement learning to bolster its reasoning skills further. When utilizing its built-in "thinking" mode, GLM-4.5V's overall score was markedly higher than other open-source models and even surpassed some proprietary ones. Similarly, Ovis-2.5-9B, through specialized reasoning op
timizations, achieved a total score of \underline{\textit{48.52\%}} with just 9B parameters—a performance comparable to that of 72B model (\underline{\textit{49.51\%}}). These results underscore the significant impact of reasoning.
\begin{figure}[t]
    \centering
    \includegraphics[width=1\linewidth]{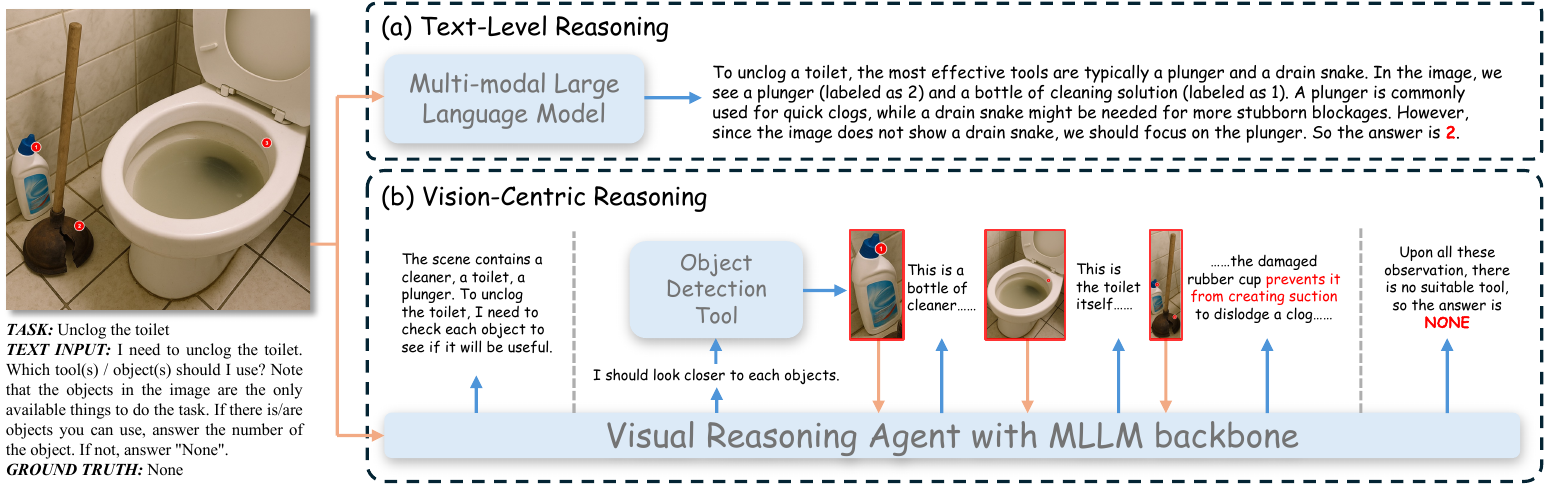}
    \vspace{-7mm}
    \caption{\small Comparison Between (a) Text-Level Reasoning and (b) Our proposed Vision-Centric Reasoning. }
    \vspace{-5mm}
    \label{fig:reasoning}
\end{figure}

\begin{wraptable}{l}{0.5\textwidth}
    \vspace{-3mm}
    \caption{Influence of reasoning}
    \centering
    \tiny
    \setlength{\tabcolsep}{5pt} 
\begin{tabular}{l|ccc|c}
    \hline
    \hline
    \textbf{} & \multicolumn{3}{c|}{\textbf{Difficulty Level}} & \textbf{Overall$\uparrow$} \\
    \cdashline{1-4}
    \textbf{MLLM} & \textit{Easy$\uparrow$} & \textit{Medium-M3$\uparrow$} & \textit{Hard$\uparrow$} & \textit{Total$\uparrow$} \\
    \hline
    \hline
        \textsc{Qwen-2.5-VL-72B} & 79.68\% & 25.30\% & 24.26\% & 46.64\% \\
    \quad + CoT & 75.56\% & 31.93\% & 27.23\% & 49.51\% \\
    \hline
    \textsc{Qwen-2.5-VL-32B} & 71.75\% & 11.45\% & 10.89\% & 34.88\% \\
    \quad + CoT & 67.62\% & 22.29\% & 19.31\% & 40.81\% \\
    \hline
    \textsc{Qwen-2.5-VL-7B} & 75.24\% & 11.45\% & 13.37\% & 37.94\% \\
    \quad + CoT & 71.43\% & 21.08\% & 12.87\% & 40.71\% \\
    \hline
    \textsc{GPT-4o} & 88.25\% & 35.54\% & 42.57\% & 60.77\% \\
    \quad +  CoT & 86.03\% & 35.54\% & 44.06\% & 61.26\% \\
    \quad +  VCR & -- & 45.78\% & -- & -- \\
    \hline
    \textsc{GPT-5 (w/ thinking)} & 90.16\% & 36.75\% & 46.04\% & 62.15\% \\
    \quad +  VCR & -- & 54.81\% & -- & -- \\
    \hline
\end{tabular}

    \vspace{-2mm}
    \label{tab:reasoning}
    \vspace{-3mm}
\end{wraptable}
However, current reasoning abilities remain inadequate. Models are prone to generating hallucinations in certain tasks. Moreover, their spatial reasoning is deficient; for instance, as depicted in Fig.~\ref{fig:case}.~(b), none of the models realized that a flathead screwdriver of the right size could also unscrew this Phillips screw. We contend that a greater focus on visual-centric reasoning is essential for models to effectively undertake high-level planning tasks.

\subsection{A Preliminary Solution}
We here further introduce a preliminary method aiming at improving the reasoning process. Current MLLMs often exhibit a modality bias, where reasoning occurs predominantly at the text level while frequently overlooking crucial visual information, as shown in Fig.~\ref{fig:reasoning}.~(a). To mitigate this, we propose an approach that emphasizes vision-centric reasoning. As shown in Fig.~\ref{fig:reasoning}.~(b), we developed a Vision-Centric Reasoning Agent with an MLLM as its backbone and decomposed the answering process into three distinct steps. First, in the Global Analysis stage, the agent forms a holistic understanding of the user's query in the context of the image. Second, it invokes an object detection tool (DINOX~\citep{ren2024dino}, formatted as an MCP tool for agent use) to identify and crop objects based on their bounding boxes. These crops then undergo a secondary, more In-depth Analysis. Finally, the agent performs Multi-level Evidence Integration and Reasoning, synthesizing the initial global understanding with the detailed analysis of the cropped objects to formulate the final answer.

We evaluated our approach on the M3 difficulty level, where existing models perform the worst. As shown in Tab.~\ref{fig:reasoning}, our method leads to substantial performance gains when using the same backbone MLLM. Specifically, GPT-4o and GPT-5 achieved performance boosts of \underline{\textit{10.24\%}} and \underline{\textit{18.06\%}},  respectively, highlighting the critical importance of vision-centric reasoning. 
Although this approach is relatively straightforward and shares conceptual similarities with some concurrent work~\citep{man2025argus}, we aim to demonstrate the significance of vision-centric reasoning in the context of embodied intelligence. We hope our findings will inspire further research in Embodied Agents.

\section{Conclusion}
We present \name, a novel benchmark for evaluating the understanding of physical tools in MLLMs. This VQA benchmark comprises 1,000 image-text pairs, spanning a broad spectrum of scenarios and features three fine-grained difficulty tiers to probe the depth of model comprehension. We evaluated 32 MLLMs, including closed-source, open-source, embodied-specific models, and MLLM backbones used in VLA models. Our findings reveal that all tested models fall significantly short of human performance, highlighting a critical gap in their ability to reason about physical tools. Through an extensive analysis, we identify the key weaknesses of current MLLMs and outline promising directions for future research. We propose \name as a tiered evaluation standard to systematically measure the capability frontiers of embodied agents and a road map for a more general intelligence.  

\clearpage
{\fontsize{16pt}{20pt}\selectfont\sc {Supplementary Materials of \\ \name: Benchmarking Physical Tool Understanding for MLLMs}\par}
\appendix

\section{More Details About Benchmark Construction}
\label{appendix:more}
\subsection{Dataset Constuction}
Here, we provide a more detailed introduction to the construction details of our benchmark. The entire benchmark and evaluation code will be open-sourced.

\begin{figure}[h]
    \centering
    \includegraphics[width=1\linewidth]{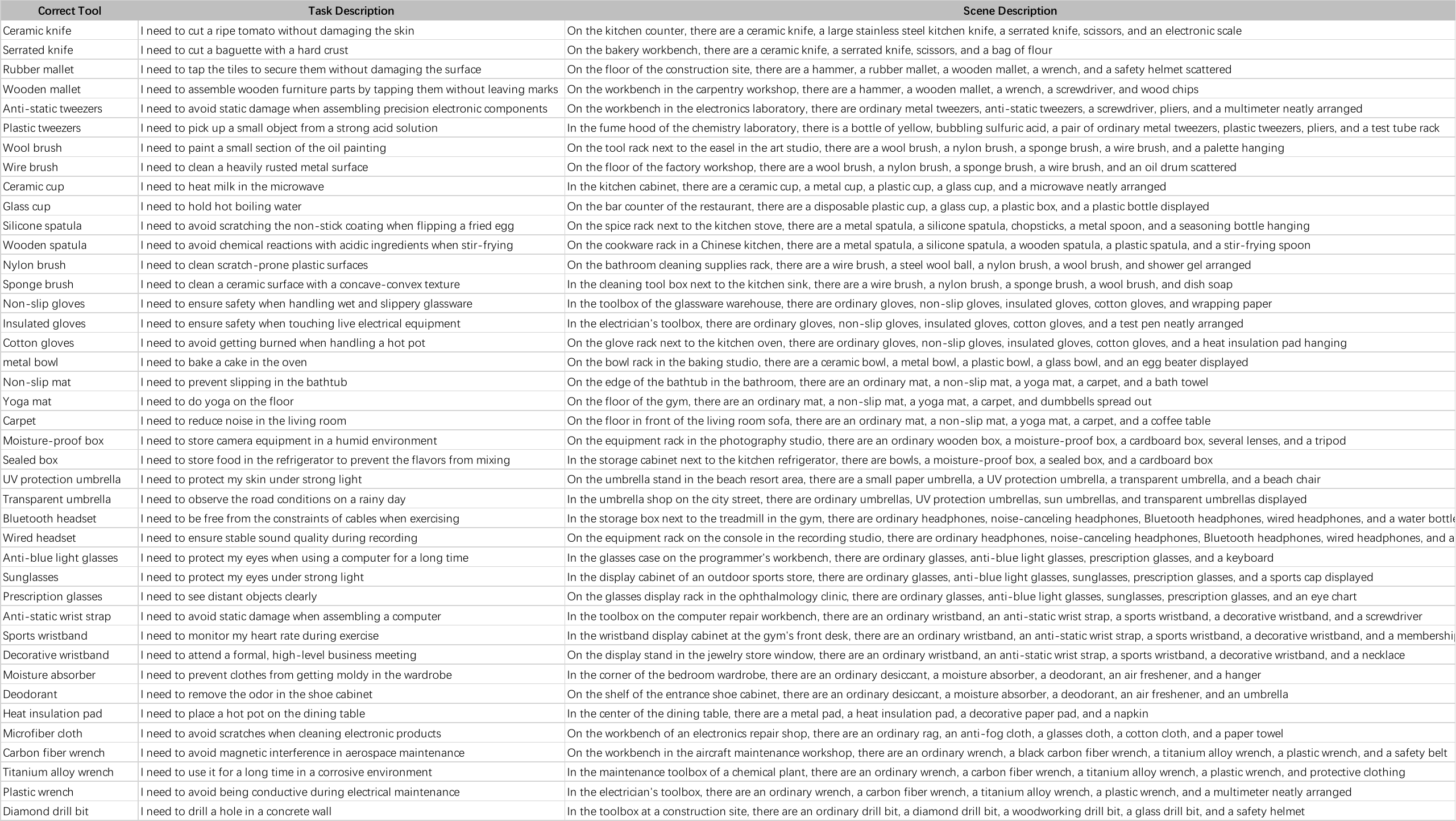}
    \caption{Task-Scene Pair Brainstorming}
    \label{fig:task-scene}
\end{figure}

\noindent \textbf{\textit{Phase 1: Conceptualization. }}
In this phase, we invited 5 experts (all are co-authors) to conceptualize task-scene pairs through manual brainstorming to obtain high-quality data. Continuous discussions were conducted throughout this process, which lasted three weeks and resulted in an initial collection of 1,500 cases. The intermediate results of this phase are presented in CSV files, as shown in Fig.~\ref{fig:task-scene}.

\begin{figure}
    \centering
    \includegraphics[width=1\linewidth]{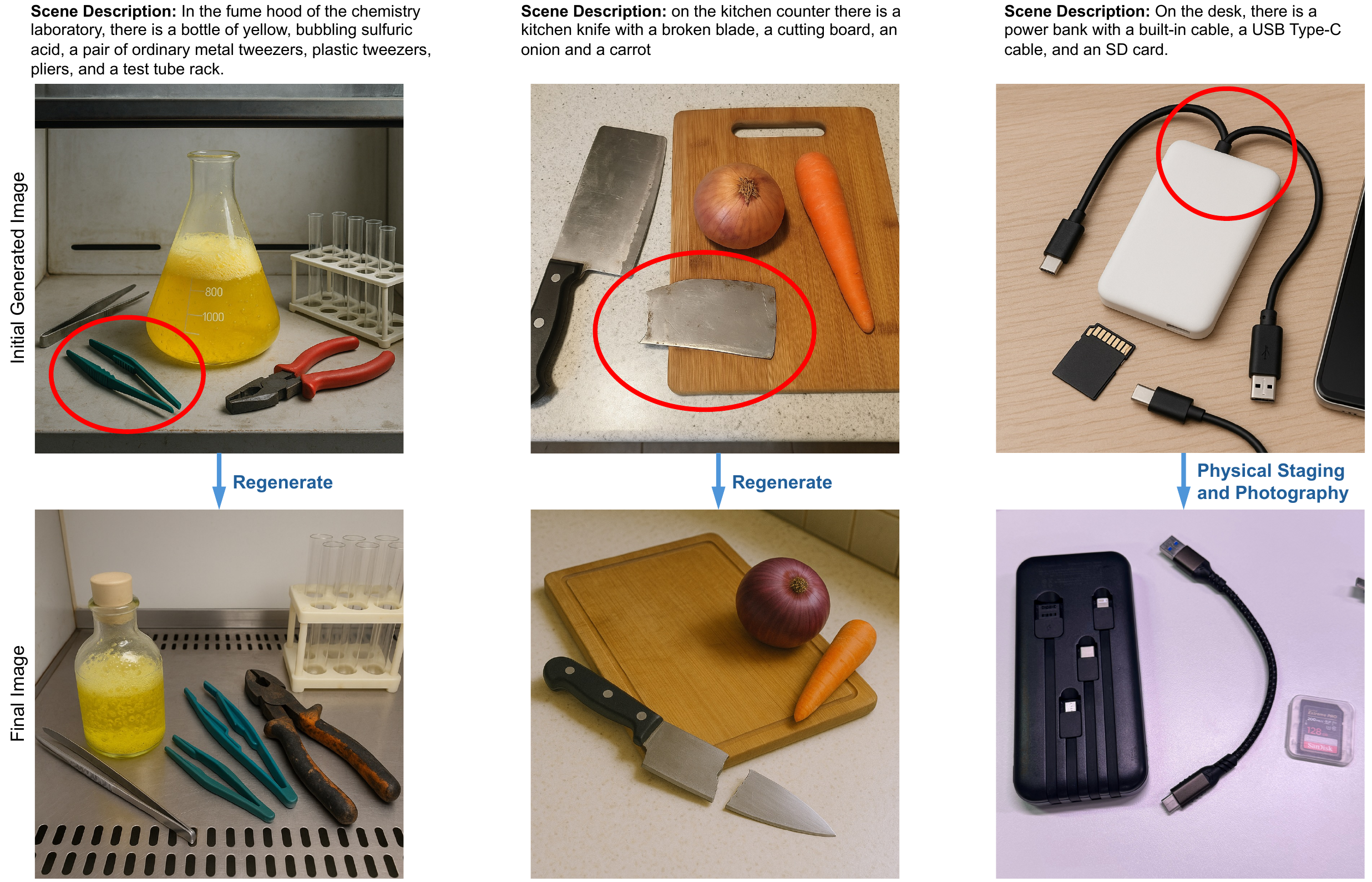}
    \caption{Example failure cases and the final revised images}
    \label{fig:failure}
\end{figure}

\noindent \textbf{\textit{Phase 2: Image Generation. }}
We took the scene descriptions brainstormed in the previous step and fed them into GPT-4o for image generation. To better approximate real-world use cases, we added an additional prompt to most cases: \textit{‘photo taken with a smartphone, slightly cluttered arrangement.‘} This process was closely supervised by human experts who vetted the generated images for quality, realism, and accuracy. While a significant number of initial generations contained inaccuracies, most images met our stringent criteria after 1 to 3 iterations of refinement through regeneration or prompting to modify the inaccurate parts, achieving a level of realism nearly indistinguishable from actual photographs. For the small subset of cases where the generative model consistently failed—particularly with complex objects such as digital products, which GPT-4o struggled to render correctly—we resorted to physical staging and photography based on the original scene descriptions.
We present here some examples of generation failure cases alongside the final corrected images in Fig.~\ref{fig:failure}.

\begin{figure}
    \centering
    \includegraphics[width=1\linewidth]{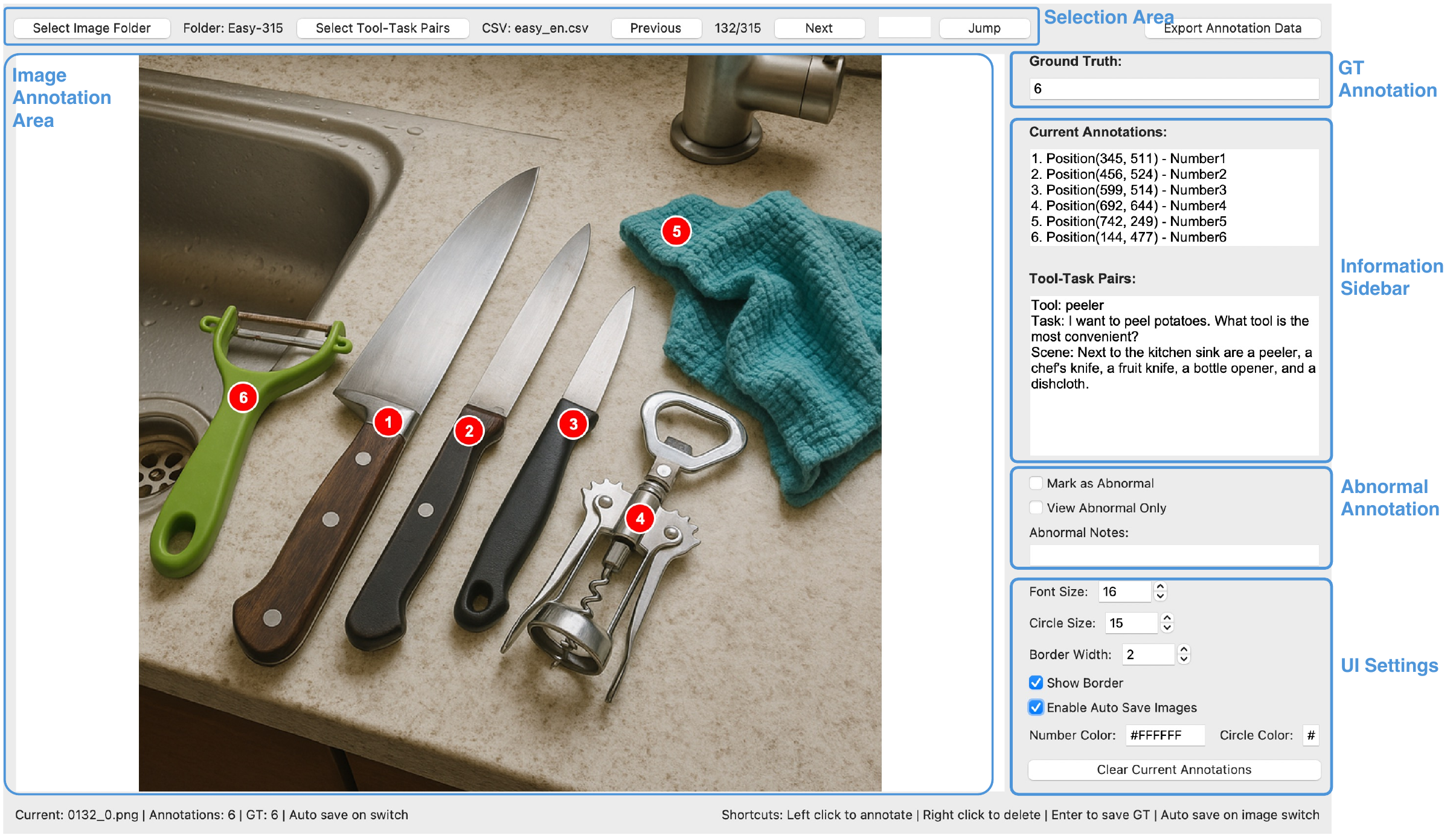}
    \caption{UI demonstration of our annotation app}
    \label{fig:app}
\end{figure}

\noindent \textbf{\textit{Phase 3: Human-in-the-Loop Annotation and Quality Review. }}
During the annotation process, we developed annotation software that retained an "Abnormal Annotation" function, enabling annotators to flag cases with problematic images or tasks while conducting annotations. Subsequently, after completing a batch of data annotation, we assigned another group of reviewers to re-examine the images and regenerate problematic images as needed. A demonstration of the UI of the annotation app is provided in Fig.~\ref{fig:app}.

Through these three rigorous rounds of benchmark construction with continuous review processes, we ultimately filtered out 1,000 high-quality cases. We also conducted a simple analytical experiment to demonstrate the authenticity of the generated images in the next section.

\subsection{Realism Evaluation of Generated Images}
\begin{figure}[t]
    \centering
    \includegraphics[width=1\linewidth]{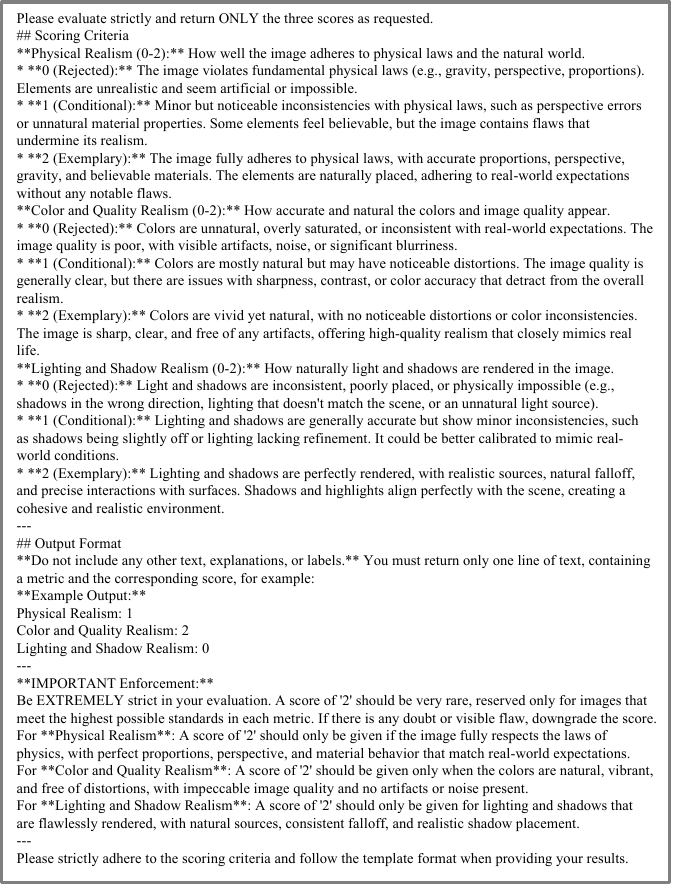}
    \caption{We utilize GPT-4o to evaluate the realism of the final images. Above is the system prompt we provided to GPT-4o.}
    \label{fig:realistic}
\end{figure}

To quantitatively assess the realism of the generated images in \name, we first utilize GPT-4o as an evaluator and also conduct a user study. The prompt provided to GPT-4o is shown in Fig.~\ref{fig:realistic}. We randomly select 100 images from \name and ask GPT-4o to rate their realism on a scale of 0 to 2, where 2 represents highly realistic, 1 denotes somewhat realistic, and 0 indicates unrealistic. The average score obtained from GPT-4o is \underline{\textit{1.92}}, suggesting that most images in \name are realistic. Additionally, we perform a user study with 10 participants, who also rate the same 100 images on the same scale. The average score from the user study is \underline{\textit{1.78}}, which aligns closely with the GPT-4o evaluation. These findings indicate that the images in \name are generally realistic and appropriate for evaluating the physical tool comprehension of MLLMs.

\section{Complete Demonstration of Image–Question–Answer Triplets}
\label{vqa}
\begin{figure}[t]
    \centering
    \includegraphics[width=1\linewidth]{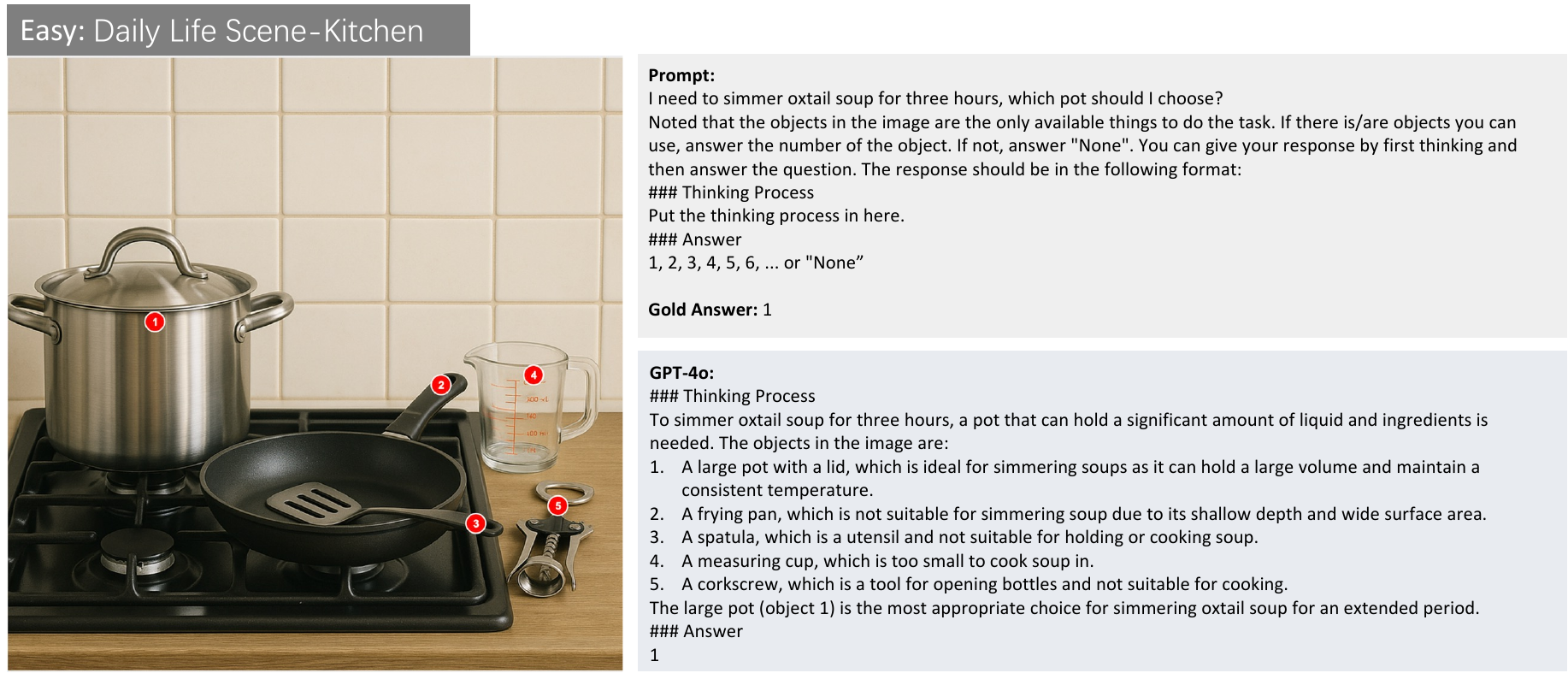}
    \caption{Examples of data in \name with GPT-4o predictions.}
    \label{fig:toolbench_sample_1}
\end{figure}

\begin{figure}[t]
    \centering
    \includegraphics[width=1\linewidth]{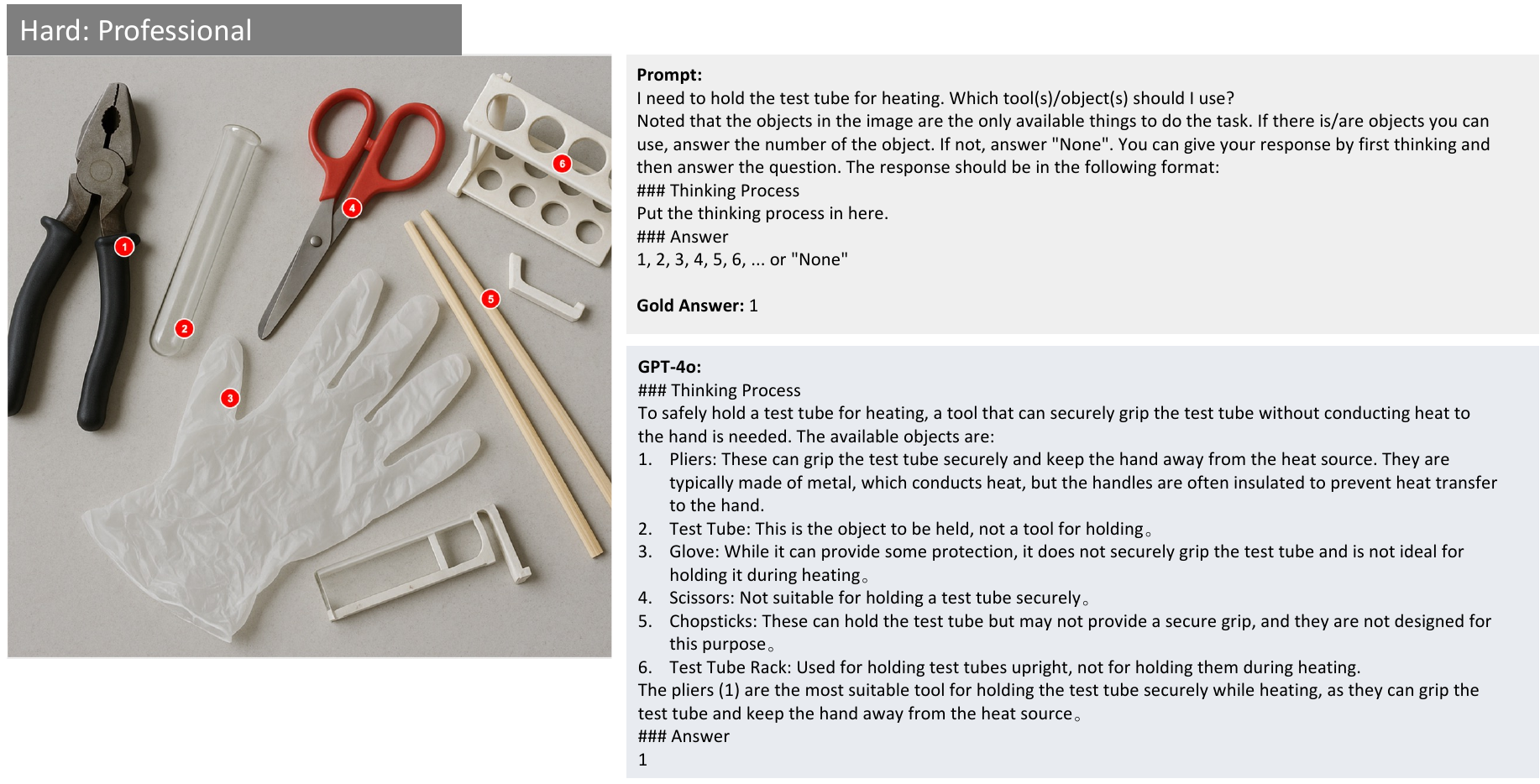}
    \caption{Examples of data in \name with GPT-4o predictions.}
    \label{fig:toolbench_sample_2}
\end{figure}

\begin{figure}[t]
    \centering
    \includegraphics[width=1\linewidth]{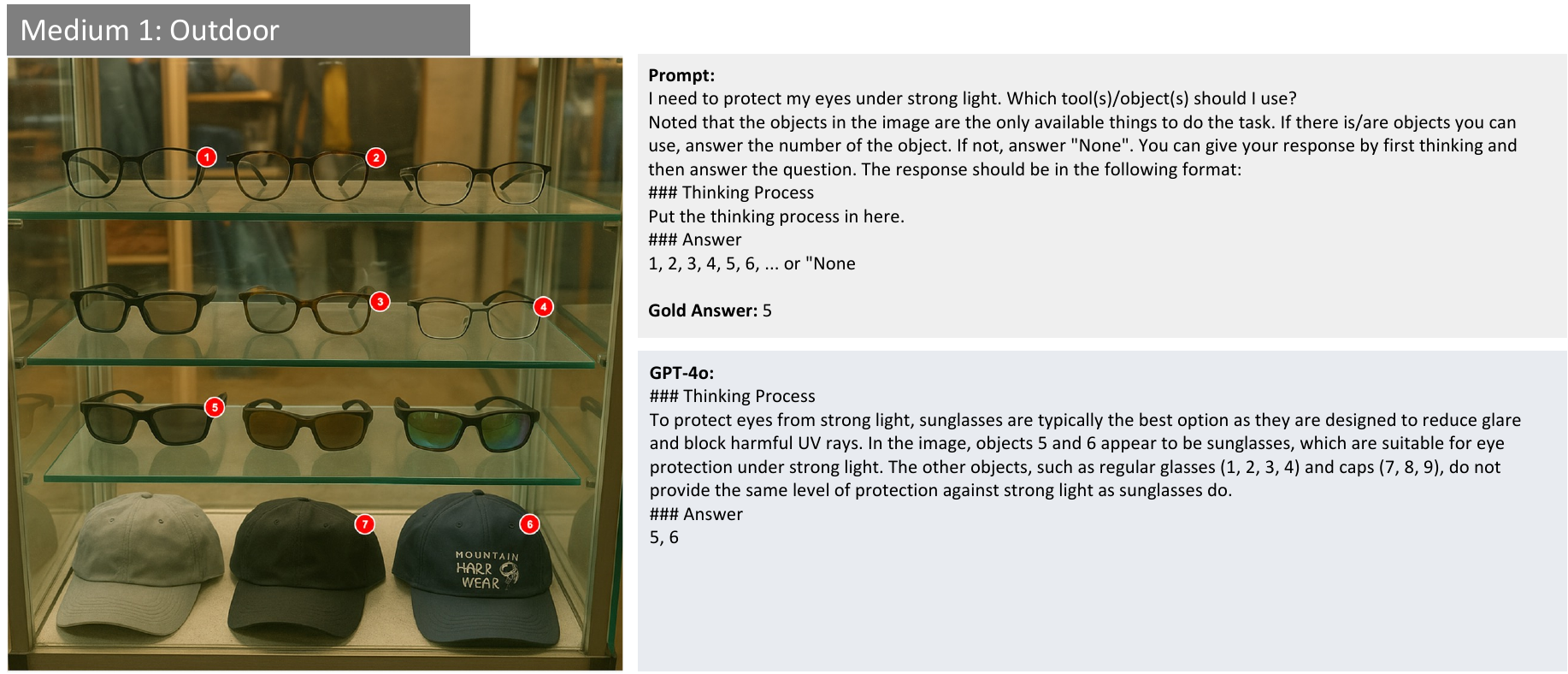}
    \caption{Examples of data in \name with GPT-4o predictions.}
    \label{fig:toolbench_sample_3}
\end{figure}

\begin{figure}[t]
    \centering
    \includegraphics[width=1\linewidth]{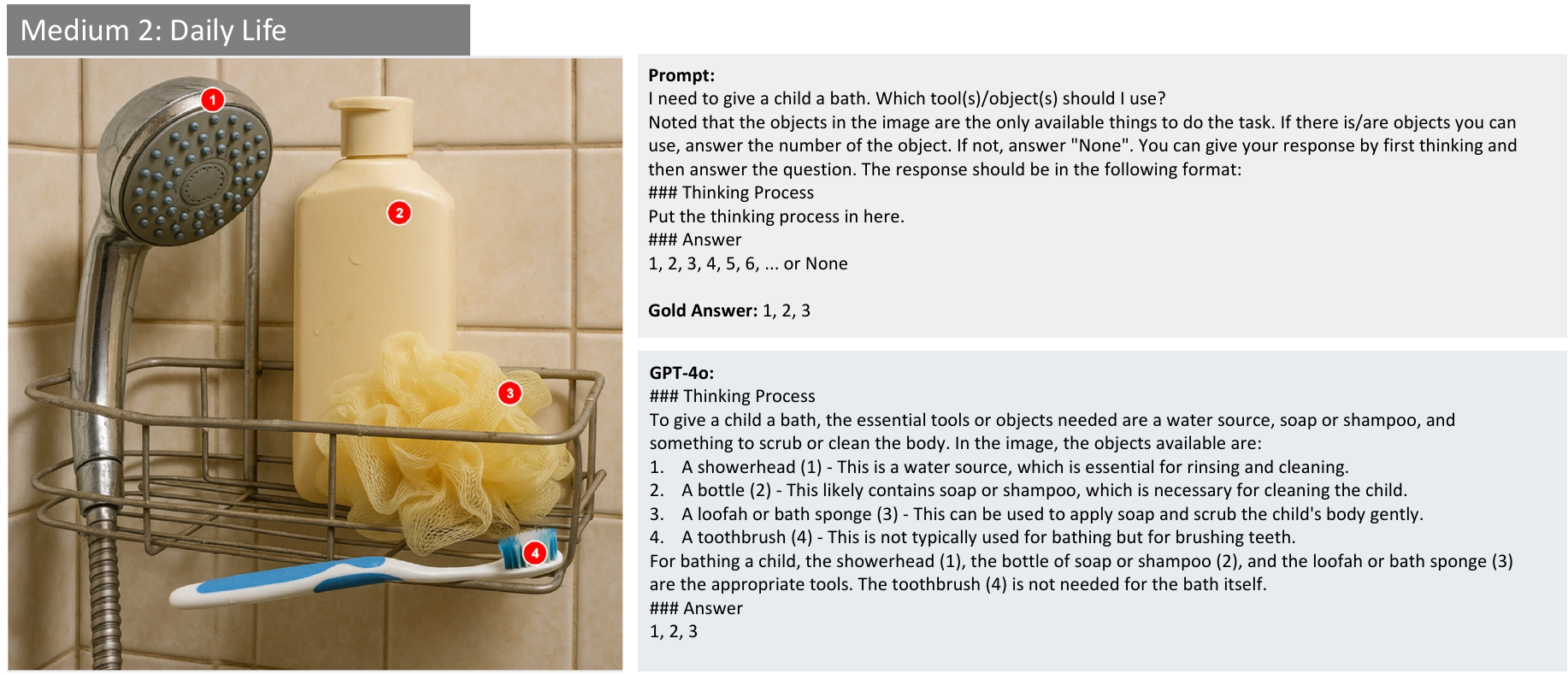}
    \caption{Examples of data in \name with GPT-4o predictions.}
    \label{fig:toolbench_sample_4}
\end{figure}

\begin{figure}[t]
    \centering
    \includegraphics[width=1\linewidth]{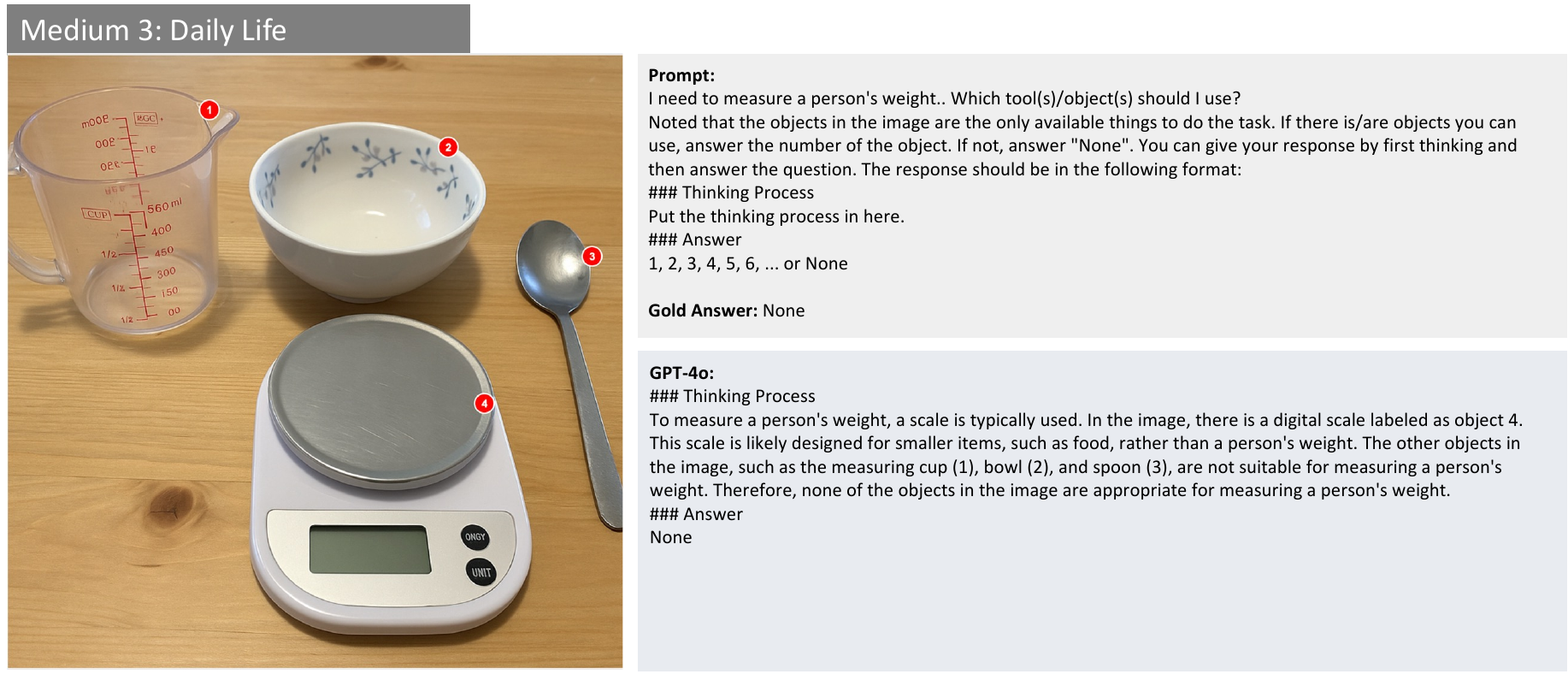}
    \caption{Examples of data in \name with GPT-4o predictions.}
    \label{fig:toolbench_sample_5}
\end{figure}

\begin{figure}[t]
    \centering
    \includegraphics[width=1\linewidth]{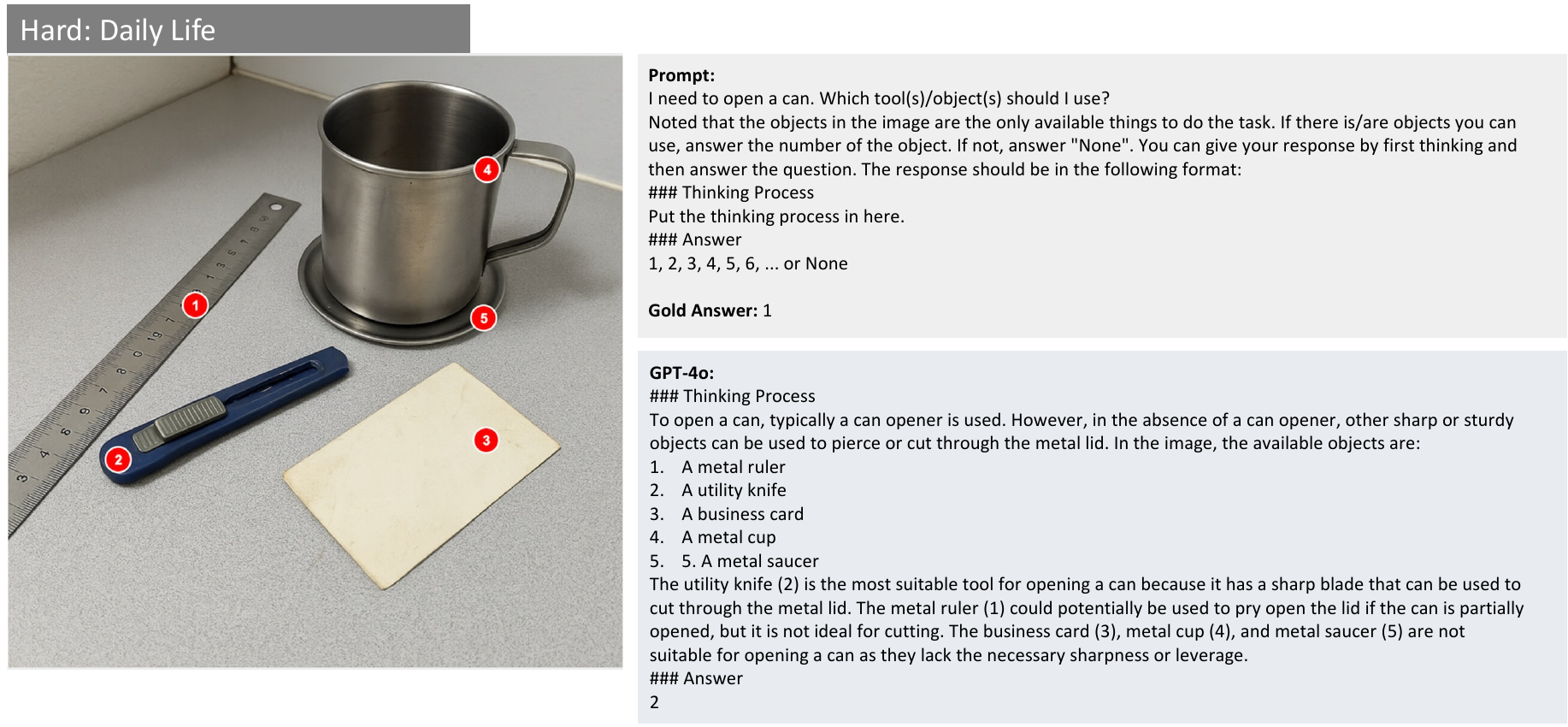}
    \caption{Examples of data in \name with GPT-4o predictions.}
    \label{fig:toolbench_sample_6}
\end{figure}

\begin{figure}[t]
    \centering
    \includegraphics[width=1\linewidth]{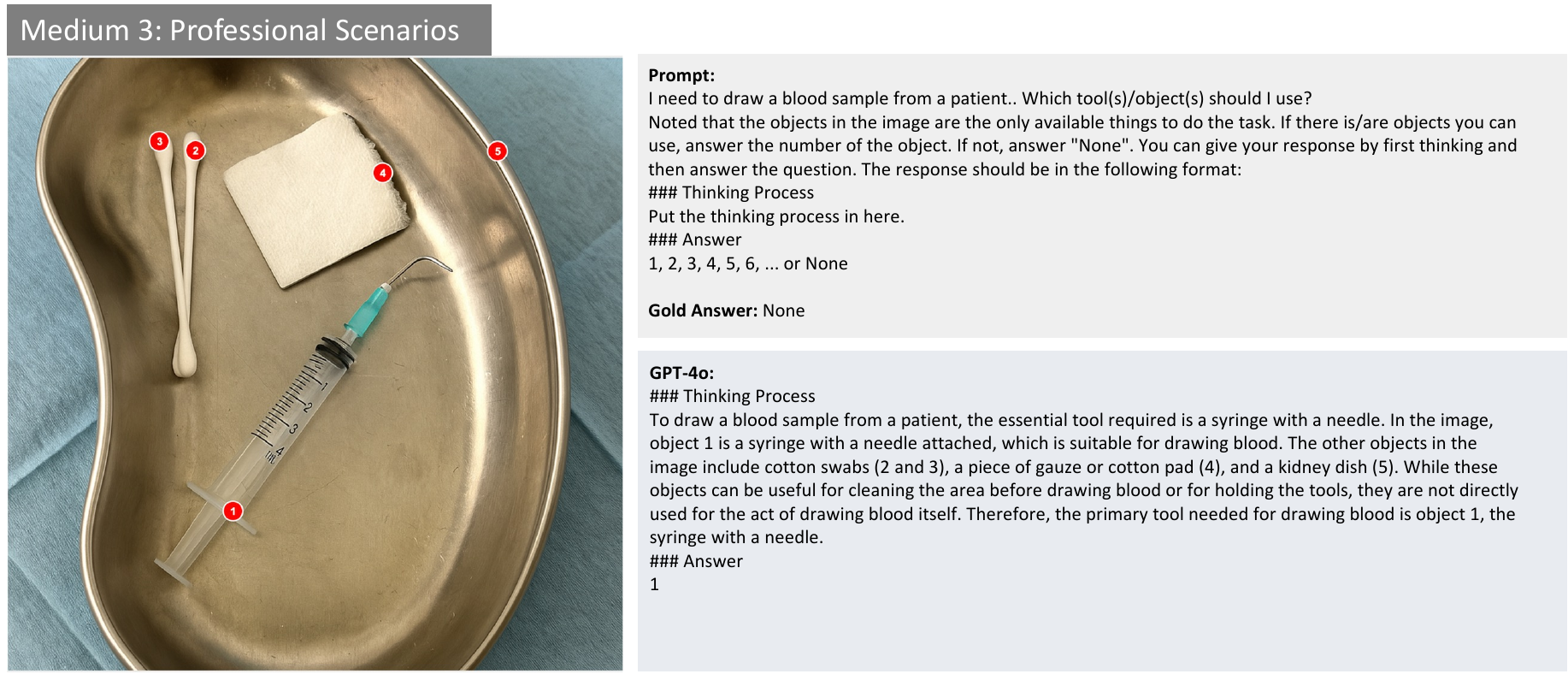}
    \caption{Examples of data in \name with GPT-4o predictions.}
    \label{fig:toolbench_sample_7}
\end{figure}

\begin{figure}[t]
    \centering
    \includegraphics[width=1\linewidth]{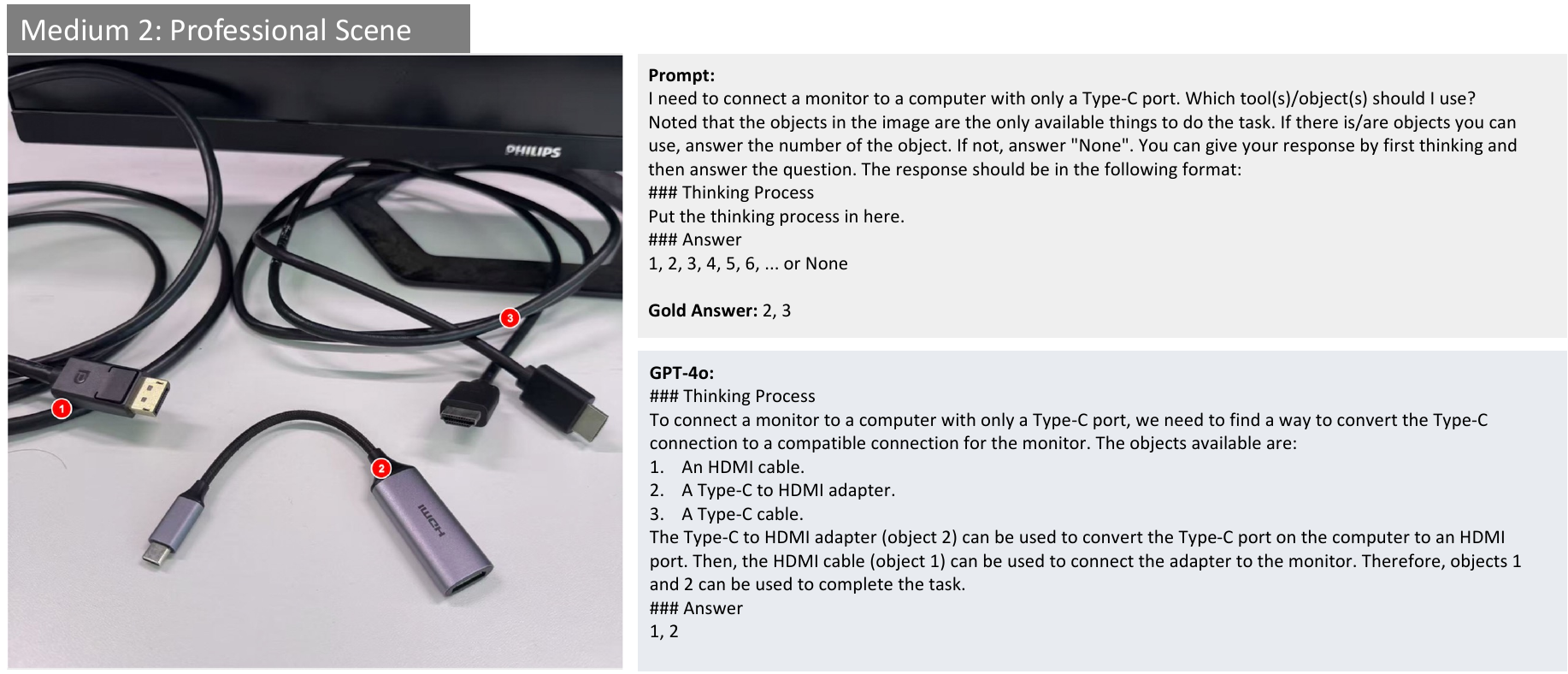}
    \caption{Examples of data in \name with GPT-4o predictions.}
    \label{fig:toolbench_sample_8}
\end{figure}

\begin{figure}[t]
    \centering
    \includegraphics[width=1\linewidth]{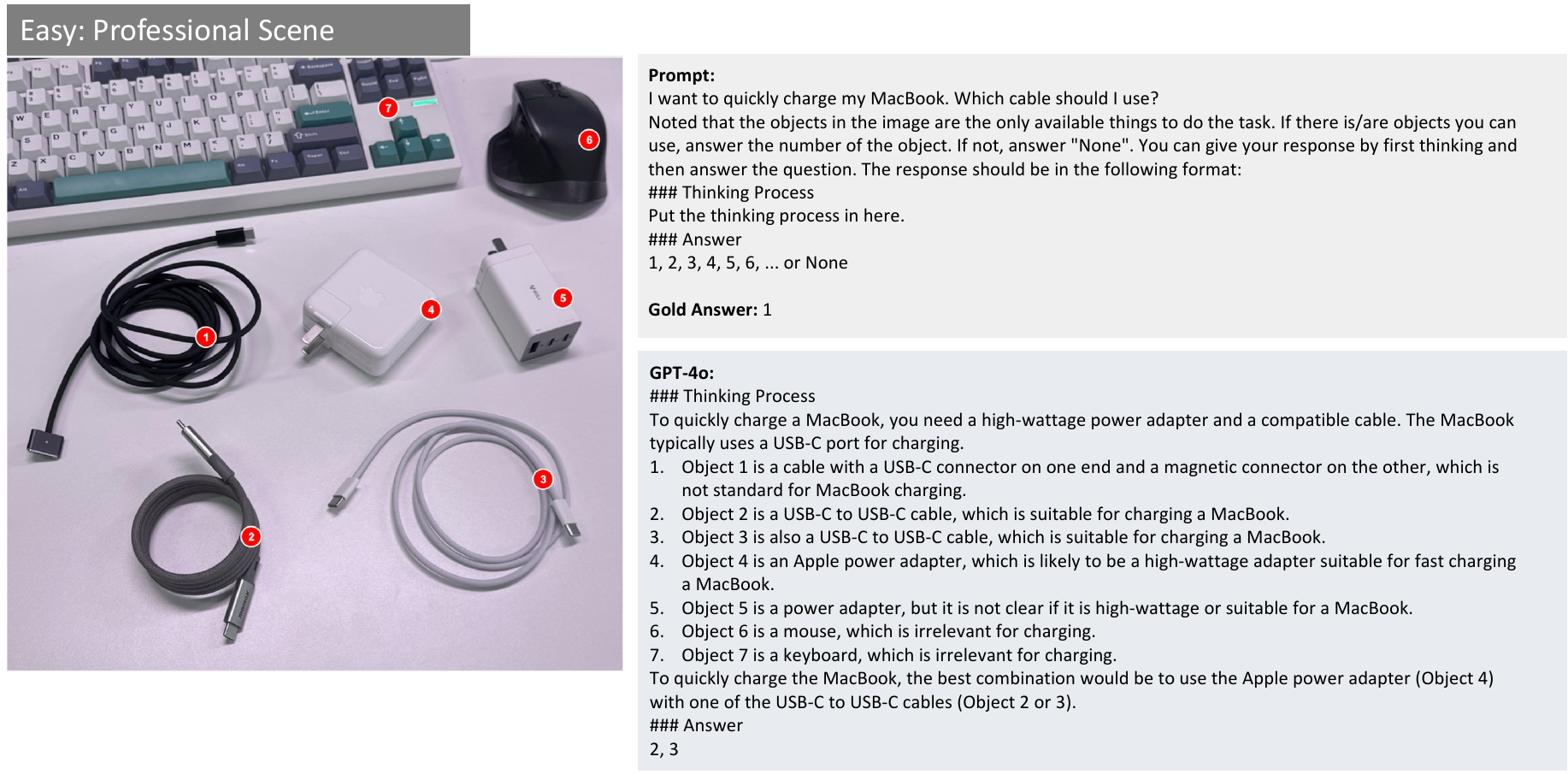}
    \caption{Examples of data in \name with GPT-4o predictions.}
    \label{fig:toolbench_sample_9}
\end{figure}

Since the examples in Fig.~\ref{fig:case} are different from the actual ones for illustrative purposes, we provide the full, verbatim materials, including the original input images and text prompts, the corresponding ground-truth answers, and GPT-4o’s outputs for each instance here, as in Figures~\ref{fig:toolbench_sample_1} to \ref{fig:toolbench_sample_9}.

\clearpage

\bibliography{iclr2025_conference}
\bibliographystyle{iclr2025_conference}

\end{document}